\documentclass[sigconf,nonacm]{acmart}

\usepackage{algorithm}
\usepackage{algorithmic}
\usepackage{multirow}
\usepackage{subfigure}
\usepackage{enumitem}
\usepackage{amsmath}

\usepackage{amssymb}
\usepackage{color}
\usepackage{balance}

\AtBeginDocument{%
  }

\begin{document}

\title{Revisiting Synthetic Human Trajectories: \\Imitative Generation and Benchmarks Beyond Datasaurus}

% \affiliation{%

\author{Bangchao Deng}
\affiliation{%
  % \institution{Department of Computer and Information Science, University of Macau}
  \institution{University of Macau}
  \city{Macao SAR}
  \country{China}}
\email{yc37980@um.edu.mo}

\author{Xin Jing}
\affiliation{%
  \institution{University of Macau}
  \city{Macao SAR}
  \country{China}}
\email{yc27431@um.edu.mo}

\author{Tianyue Yang}
\affiliation{%
  \institution{University of Macau}
  \city{Macao SAR}
  \country{China}}
\email{mc35301@um.edu.mo} 

\author{Bingqing Qu}
\affiliation{%
  \institution{BNU-HKBU United International College, China}
  \city{}
  \country{}}
\email{bingqingqu@uic.edu.cn}

\author{Dingqi Yang}
\authornote{Corresponding author}
\affiliation{%
  \institution{University of Macau}
  \city{Macao SAR}
  \country{China}}
\email{dingqiyang@um.edu.mo}

\author{Philippe Cudre-Mauroux}
\affiliation{%
  \institution{University of Fribourg}
  \city{}
  \country{Switzerland}}
\email{philippe.cudre-mauroux@unifr.ch}

% \affiliation{%

% \renewcommand{\shortauthors}{Bangchao Deng et al.}

\renewcommand{\shortauthors}{Bangchao Deng, Xin Jing, Tianyue Yang, Bingqing Qu, Dingqi Yang, and Philippe Cudre-Mauroux}
% Bangchao Deng, Xin Jing, Tianyue Yang, Bingqing Qu, Dingqi Yang, and Philippe Cudre-Mauroux
\begin{abstract}
Human trajectory data, which plays a crucial role in various applications such as crowd management and epidemic prevention, is challenging to obtain due to practical constraints and privacy concerns. In this context, synthetic human trajectory data is generated to simulate as close as possible to real-world human trajectories, often under summary statistics and distributional similarities. However, these similarities oversimplify complex human mobility patterns (a.k.a. ``Datasaurus''), resulting in intrinsic biases in both generative model design and benchmarks of the generated trajectories. Against this background, we propose MIRAGE, a hu\underline{M}an-\underline{I}mitative t\underline{RA}jectory \underline{G}en\underline{E}rative model designed as a neural Temporal Point Process integrating an Exploration and Preferential Return model. It imitates the human decision-making process in trajectory generation, rather than fitting any specific statistical distributions as traditional methods do, thus avoiding the Datasaurus issue. We also propose a comprehensive task-based evaluation protocol beyond Datasaurus to systematically benchmark trajectory generative models on four typical downstream tasks, integrating multiple techniques and evaluation metrics for each task, to assess the ultimate utility of the generated trajectories. We conduct a thorough evaluation of MIRAGE on three real-world user trajectory datasets against a sizeable collection of baselines. Results show that compared to the best baselines, MIRAGE-generated trajectory data not only achieves the best statistical and distributional similarities with 59.0-67.7\% improvement, but also yields the best performance in the task-based evaluation with 10.9-33.4\% improvement. A series of ablation studies also validate the key design choices of MIRAGE. Our code and datasets are at \url{https://github.com/UM-Data-Intelligence-Lab/MIRAGE}.
\end{abstract}
% Our code and datasets are in https://anonymous.4open.science/r/M-C.

% , and outperform best baselines by 3.46-10.65\% across different tasks.


% \ccsdesc[500]{Computing methodologies~Modeling and simulation}



% \settopmatter{printacmref=true}
\maketitle

\section{Introduction}
% TrajGen \cite{cao2021generating} takes real
% mobility trajectories and the corresponding mobility map as inputs (mobility map can be denoted as a graph G(E,V), where E refers to the set of edges that provide connectivity (e.g., road segments for vehicles or pedestrians), and V refers to intersections that provide transitions (e.g., road junctions).)

% TS-TrajGen \cite{jiang2023continuous}  each segment pair
% (xi, xi+1) is adjacent in the road network graph. current methods are not continuous routes on the road network, which makes these synthetic trajectories unusable for downstream applications like traffic simulation.

Human trajectory data is a key ingredient for a wide range of applications, including urban planning \cite{urban}, traffic management \cite{traffic}, epidemic analysis \cite{epidemic}, predictive policing \cite{yang2018crimetelescope}, and crowd monitoring \cite{chen2024animating}. These applications heavily rely on the quality of human mobility models learnt from human trajectory data. However, acquiring large-scale real human trajectory data is often challenging due to practical constraints and privacy concerns. Therefore, synthetic human trajectory data has been widely used as an alternative, where generative models are learnt to generate artificial trajectories that closely resemble real-world human trajectories, making human trajectory data more readily available.

%  \cite{commercial}
%, cost-effective, and privacy-preserving.

%  summary statistics and distributional similarities -> oversimplification -> ultimate goal -> agnostic design -> imitative generation and evaluation beyond datasaurus.

% The goal of generative models of individual human mobility is to generate synthetic trajectories with realistic mobility patterns.

In the current literature, existing works on synthetic human trajectory generation mostly focus on the resemblance under summary statistics and distributional similarities \cite{kapp2023generative}. For example, these works measure the resemblance in different aspects of mobility characteristics such as spatial distribution (e.g. G-Rank) \cite{jiang2023continuous}, temporal distribution (e.g. stay duration) \cite{pappalardo2018data}, OD flows (trips per OD pair) \cite{yuan2011driving}, and user mobility patterns (e.g. I-Rank and DailyLoc) \cite{xu2021simulating}, using divergence/distance metrics such as Kullback-Leibler divergence (KLD) \cite{bindschaedler2016synthesizing}, Jensen-Shannon divergence (JSD) \cite{ouyang2018non,feng2020learning}, earth mover’s distance (EMD) \cite{anda2021synthesising}, Root Mean Squared Error (RMSE) \cite{pappalardo2018data}. These similarities often serve on one hand as part of the model design such as the model fitting objective \cite{jiang2016timegeo}, while on the other hand also as the benchmarks for evaluating trajectory generative models \cite{long2023practical,ouyang2018non,huang2019variational,xiong2023trajsgan}. However, while these similarities provide insights into the differences between real and generated data from various perspectives, they indeed oversimplify the complexity of human mobility patterns, resulting in intrinsic biases in both generative model design and benchmarks of the generated trajectories. Specifically, datasets that are similar over a number of statistical properties may yield very different patterns, known as ``Datasaurus'' \cite{cairo2016download, matejka2017same}. In the context of trajectory data, it implies that statistically/distributionally similar trajectories may imply different mobility patterns, which thus leads to different performances in downstream tasks (as evidenced by our experiments in Section \ref{sec:exp_datasaurus}). Therefore, the resemblance under summary statistics and distributional similarities cannot fully reflect the ultimate utility of generated trajectories in supporting downstream tasks. 

A few recent works started to consider the utility of downstream tasks, by either incorporating task-specific prior knowledge in the generative model design, such as constraining the next locations in a close neighborhood for traffic flow simulation \cite{jiang2023continuous}, or benchmarking generative models on a heuristically designed downstream task with one specific technique to solve the task \cite{feng2020learning, yuan2022activity, long2023practical}. Nevertheless, these works are either limited by their task-specific design or lack a comprehensive view of utility benchmarks. In particular, heuristically designed downstream tasks may lead to unknown biases in the utility evaluation, as the performance of different techniques solving the same task often varies (as evidenced by our experiments in Appendix \ref{sec:app_task_bias}) and heuristically choosing one technique as the benchmark is thus untrustworthy.

%such as transition probability \cite{bindschaedler2016synthesizing} and the distribution of time intervals between locations \cite{jiang2016timegeo}

Against this background, we study in this paper the problem of task-agnostic human trajectory generation and its benchmarks beyond ``Datasaurus''. Specifically, the task-agnostic generation should be independent of specific downstream tasks and imitate the human decision-making process in trajectory generation. Such human-imitative design needs to consider the unique characteristics of human trajectories. First, real-world human trajectories usually consist of sparse and irregularly observed presence events, where individuals voluntarily share their presence at semantic-enriched locations under their own preference, such as check-ins at Points of Interest (POIs\footnote{Given the context of human trajectories, we do not distinguish the two terms ``locations'' and ``POIs'' throughout this paper.}) on social media \cite{yang2019revisiting}, which differ from other periodic and regularly sampled mobility traces with only GPS coordinates such as taxi trajectories \cite{cao2021generating}. To handle such human trajectories, some works simply treat human trajectories as sequences without considering the temporal information \cite{he2015dpt,huang2019variational}, while some others heuristically transform human trajectories to sequences of regularly observed events under a pre-defined and dataset-specific time interval (e.g., every 30 minutes \cite{feng2020learning} or 1-hour \cite{pappalardo2018data,bindschaedler2016synthesizing}). However, these approaches are limited either in ignoring the important temporal information, or in incorporating a strong assumption of observing human trajectory in regular time intervals and heuristically interpolating missing and removing redundant events. Second, human decision-making on mobility choices has been widely recognized and evidenced to follow an Exploration and Preferential Return (EPR) model \cite{song2010modelling, jiang2016timegeo, pappalardo2018data}. It presents individuals with a choice between two competing mechanisms: the exploration mechanism selecting previously unvisited locations and the preferential return mechanism to encourage returning to a previously visited location. Existing works mostly design mechanistic EPR-alike models reliant on simple statistic models under Markovian assumptions \cite{bindschaedler2016synthesizing}. However, despite the advantage of being interpretable by design, these statistic models have intrinsic limitations in accurately modeling the complex mobility patterns observed in real-world human trajectories.

To address the above issues, we propose MIRAGE, a hu\underline{M}an-\underline{I}mitative t\underline{RA}jectory \underline{G}en\underline{E}rative model, which imitates the holistic human decision-making process in trajectory generation and does not explicitly fit any specific statistical distributions (while traditional EPR model does), thus avoiding the ``Datasaurus'' issue. Specifically, a human trajectory consisting of a sequence of stochastic presence events on continuous time is naturally a Temporal Point Process (TPP). Subsequently, we design MIRAGE as a neural TPP with intensity-free parameterization, benefiting from both the flexibility of the neural network encoding the trajectory history and the efficiency of the TPP in modeling continuous-time stochastic events \cite{shchur2021neural}. To generate one event of a human trajectory, we first sample an event time from the neural TPP based on the encoded trajectory history, and then sample the activity category further conditioned on the sampled event time. Afterward, we design a neural EPR model, mimicking the human decision-making process to choose either exploring unvisited locations or returning to previously visited locations, conditioned on the sampled time and activity category. We also adopt a user variational autoencoder to imitate individual preferences in the trajectory generation process.

Meanwhile, we also propose a comprehensive task-based evaluation protocol to systematically benchmark trajectory generative models beyond summary statistics and distributional similarities (i.e. ``Datasaurus''). The key idea is to evaluate whether the generated trajectories are similar to real trajectories in practically supporting different downstream tasks. To this end, we evaluate the performance of both real and generated mobility trajectories on a variety of typical downstream tasks using multiple techniques and evaluation metrics for each task, so as to average out the biases of individual techniques and metrics. We measure the paired performance discrepancy between the real and generated mobility trajectories using relative errors, such as Mean Absolute Percentage Error (MAPE) and Mean Squared Percentage Error (MSPE), which finally serve as benchmarks to assess the ultimate utility of the generated trajectories. 

We summarize our contribution as follows:

%where the TPP is parameterized by a recurrent neural network, 
\begin{itemize}[leftmargin=*]
    \item We reveal the limitations of existing human trajectory generative models in focusing on the summary statistics and distributional similarities between real and generated trajectories, which lead to intrinsic biases in both generative model design and benchmarks.
    %of the generated trajectories. 
    
    \item We propose MIRAGE, a hu\underline{M}an-\underline{I}mitative t\underline{RA}jectory \underline{G}en\underline{E}rative model designed as an intensity-free neural Temporal Point Process integrating a neural Exploration and Preferential Return model to imitate the human decision-making process in trajectory generation.
    
    \item We propose a comprehensive task-based evaluation protocol to systematically benchmark trajectory generative models on four typical downstream tasks, integrating multiple techniques and evaluation metrics for each task, to assess the ultimate utility of the generated trajectories.
    
    \item We conduct a thorough evaluation of MIRAGE on three real-world human trajectory datasets against a sizeable collection of state-of-the-art baselines. Results show that compared to the best baselines, MIRAGE-generated trajectory data not only achieves the best statistical and distributional similarities with 59.0-67.7\% improvement but also yields the best performance in the task-based evaluation with 10.9-33.4\% improvement.
    
\end{itemize}

\section{Related Work}
\subsection{Mobility Trajectory Generation}
%Mobility trajectory generation requires capturing the spatiotemporal patterns of individuals so as to generate similar trajectories. 
Early works mostly model individual mobility with explicit physical meanings \cite{gonzalez2008understanding, song2010limits} and generate synthetic trajectories under the distributions of key characteristics observed in real mobility patterns, such as trip lengths, start locations, or start times, etc. As a widely recognized mobility model, the Exploration and Preferential Return model (EPR) \cite{song2010modelling} unifies exploration and return mobility patterns by selecting new locations based on a random walk process for exploration and revisiting previously visited locations based on their frequency for preferential return. Subsequent studies further extend the EPR model by integrating sophisticated spatial or social information, such as mining the correlation between location capacity and social network sizes \cite{alessandretti2018evidence}, incorporating a nested gravity model into the EPR model \cite{pappalardo2015returners}, linking mobility to social ties and studying the dynamics of spatial choices based on social behavior \cite{toole2015coupling}, and integrating the circadian propensity of human mobility and Markov-based models into the EPR model \cite{jiang2016timegeo}. In addition, the EPR model is also shown to be universal in human behavior modeling in general, such as user activities in recommendation systems \cite{onuma2009tangent} and human behaviors in cyberspace \cite{hu2019return}. However, these models, often reliant on heuristic statistical assumptions, have limitations in accurately modeling the complex mobility patterns observed in real-world trajectories. 

% recommendation system \cite{sun2020framework,onuma2009tangent}
% \cite{alessandretti2018evidence, pappalardo2015returners, toole2015coupling}.
% \cite{alessandretti2018evidence} mining the correlation between location capacity and social network size. 
% \cite{pappalardo2015returners} incorporating a gravity model into the EPR mechanism.
% \cite{toole2015coupling} linking mobility to social ties and studying the dynamics of spatial choices based on social behavior.

Recently, deep learning generative models have been widely adopted for mobility trajectory generation. They can flexibly capture the complex spatiotemporal patterns encoded in real-world mobility trajectories without strong prior assumptions. For example, SeqGAN \cite{yu2017seqgan} is the pioneering work of sequence generation based on Generative Adversarial Networks (GAN); MoveSim \cite{feng2020learning} later incorporates information about physical distance, temporal periodicity, and historical transition matrix of location into a GAN framework; TrajGen \cite{cao2021generating} employs a CNN-based GAN to map mobility trajectories to images and to generate synthetic trajectory images, followed by a Seq2Seq model to output the synthetic trajectory; DeltaGAN \cite{xu2021simulating} adopts a two-stage generative model to simulate human mobility trajectories, capturing fine-grained timestamps and effectively representing temporal irregularities; TS-TrajGen \cite{jiang2023continuous} combines the A* algorithm \cite{hart1968formal} with a GAN framework to generate continuous trajectories on urban road networks; SAVE \cite{huang2019variational} combines VAE and LSTM for mobility trajectory generation.

%  and have shown promising results on the statistical resemblance between the generated and real trajectories

% , benefiting from both the ability of VAE to construct a latent space capturing essential mobility features and the efficiency of LSTM to capture sequential information. 

In addition, (neural) temporal point processes \cite{chen2020nstpp, shchur2019intensity, zuo2020transformer, yuan2023dstpp} are also widely used to model the temporal dynamics of user behaviors. In the context of trajectory generation, VOLUNTEER \cite{long2023practical} incorporates a two-layer VAE model with a temporal point process to capture the characteristics of human mobility; ActSTD \cite{yuan2022activity} enhances the dynamic modeling of individual trajectories by utilizing neural ordinary equations in the continuous location domain; DSTPP \cite{yuan2023dstpp} further models the complex spatiotemporal joint distributions using diffusion models. In this paper, beyond traditional generative deep learning models, we further design and integrate a neural EPR model with neural TPPs to imitate the human decision-making process in trajectory generation.

%However, all of the models mentioned above simplified or neglected the important and complex explore-return pattern of human mobility.

%Neural STPP \cite{chen2020nstpp} parameterizes the spatial probability density function and temporal intensity with continuous-time normalizing flows in the continuous domain. Other deep learning based methods such as neural temporal point process \cite{shchur2019intensity,zuo2020transformer} and diffusion model \cite{yuan2023dstpp} have also been employed to model the spatiotemporal event data.

% which paved the way for further advancements in sequence generation using GAN-based techniques

\subsection{Synthetic Mobility Trajectory Benchmarks}
The benchmarks for synthetic mobility trajectories can be classified into two categories \cite{kapp2023generative}. First, statistical and distributional similarities are the most widely used benchmarks, such as Kullback-Leibler divergence (KLD) \cite{bindschaedler2016synthesizing}, Jensen-Shannon divergence (JSD) \cite{ouyang2018non,feng2020learning}, earth mover’s distance (EMD) \cite{anda2021synthesising}, Root Mean Squared Error (RMSE) \cite{pappalardo2018data}, which are used to measure the similarities between real and generated trajectories in different aspects. For example, typical mobility statistics include the radius of gyration \cite{feng2020learning,jiang2023continuous,yuan2022activity}, the number of distinct locations visited per user per day \cite{feng2020learning,xu2021simulating}, I-Rank (frequency of visiting personal top locations) \cite{feng2020learning,pappalardo2018data}, and the number of daily trips per user \cite{pappalardo2018data}, trip lengths between consecutive trajectory points or between the origin and destination \cite{feng2020learning, jiang2023continuous, xiong2023trajsgan}; spatial distributions characterize the distribution of locations based on factors like visits per location (i.e. G-Rank) \cite{feng2020learning, jiang2023continuous, ouyang2018non, wang2021large, xiong2023trajsgan} or location popularity ranking \cite{du2023ldptrace}; temporal distributions characterize the number of trips per hour of the day \cite{pappalardo2018data}, stay duration \cite{feng2020learning,long2023practical,ouyang2018non} and time intervals between check-ins \cite{yuan2022activity}. However, while these similarity metrics provide insights into the differences between real and generated data from various perspectives, they cannot fully reflect the ultimate utility of generated trajectories in supporting downstream tasks.

% : 1) the statistical and distributional similarity between real and generated trajectories, and 2) the utility in supporting downstream tasks
% but are not limited to 

Second, benchmarking on downstream tasks recently emerged as an evaluation scheme for synthetic mobility trajectories. These tasks include road map updating \cite{cao2021generating}, next location prediction \cite{jiang2023continuous,long2023practical}, and spreading simulation \cite{xiong2023trajsgan,feng2020learning,yuan2022activity}. However, these works often use a heuristically designed downstream task with one specific technique/algorithm to solve the task, which leads to unknown biases in the utility evaluation. As evidenced in our experiments in Appendix \ref{sec:app_task_bias}, the performance of different techniques solving the same task varies; heuristically choosing the results of one technique as the benchmark is thus untrustworthy. Therefore, we propose a comprehensive task-based evaluation protocol to systematically benchmark synthetic mobility trajectories.

\section{Preliminaries}
\subsection{Problem Definition}
\noindent \textbf{Human Trajectory.} A trajectory is defined as a time-ordered sequence $X$ = \{$x_1$, $x_2$, ..., $x_n$\}, where $x_i =(t_i, k_i, l_i)$ is a presence event defined as a tuple consisting of a timestamp $t_i$, and a semantic activity category $k_i$, and a location (POI) $l_i$.

\noindent \textbf{Trajectory Generation.} Given a real-world human trajectory dataset, the objective is to generate a new trajectory dataset while preserving the fidelity and utility of the original real-world dataset.

\subsection{Neural Temporal Point Processes}
\subsubsection{Temporal Point Processes}
A Temporal Point Process (TPP) is a stochastic process where its realization is a sequence of discrete events in time, represented as a sequence $\mathcal{T} = \{t_1, ..., t_N\}$, which can be equivalently represented as a sequence of strictly positive inter-event times $\tau_{i+1} = (t_{i+1}-t_i) \in \mathbb{R}^+$. The conditional intensity $\lambda^*(t)$, which fully specifies the TPP, represents the instantaneous rate of arrival of new events at time $t$ given the history of past events $\mathcal{H}_{t}$ = $\{t_j \in \mathcal{T} |t_j < t\}$, where (*) is used as a shorthand for conditioning on the history. The conditional probability density function of time $\tau_{i+1}$ until the next event is computed by the integration of $\lambda^*(t)$ as follows:
\begin{equation}
p^*(\tau_{i+1}) = {\lambda}^{*}(t_{i}+\tau_{i+1}) \exp \left( - \int_0^{\tau_{i+1}} \lambda^{*}(t_{i}+s)ds \right) 
\label{TPP_PDF}
\end{equation}

\subsubsection{TPP Parameterization}
Traditional TPPs often specify a simple parametric intensity function capturing relatively simple patterns in event occurrences, such as the Hawkes process \cite{hawkes1971spectra}, which often leads to poor results because of its limited flexibility in modeling complex data. In this context, neural TPPs are developed to use neural networks to learn complex dependencies from the history of TPPs, and then approximate the intensity function by some parametric forms \cite{shchur2021neural}. However, these intensity-based approaches cannot achieve flexibility, efficiency, and ease-to-use simultaneously \cite{shchur2019intensity}. Alternatively, an intensity-free learning model of neural TPPs is proposed to use a mixture of log-normal distributions to represent the conditional probability density function $p^*(\tau)$ directly, and thus bypass the cumbersome intensity function \cite{shchur2019intensity}. Due to its advantage of closed-form sampling and likelihood computation, the intensity-free approach has been widely adopted in learning neural TPPs \cite{xu2021simulating, feng2023memory}. We also adopt this approach in our work.

% gupta2022modeling, 

\section{MIRAGE}
To imitate the human decision-making process in trajectory generation, we design MIRAGE as an intensity-free neural Temporal Point Process (TPP) integrating a neural Exploration and Preferential Return (EPR) model. In the following, we first present the overview of MIRAGE, followed by the details of individual components.

\begin{figure}
    \centering
    \includegraphics[width=0.48\textwidth]{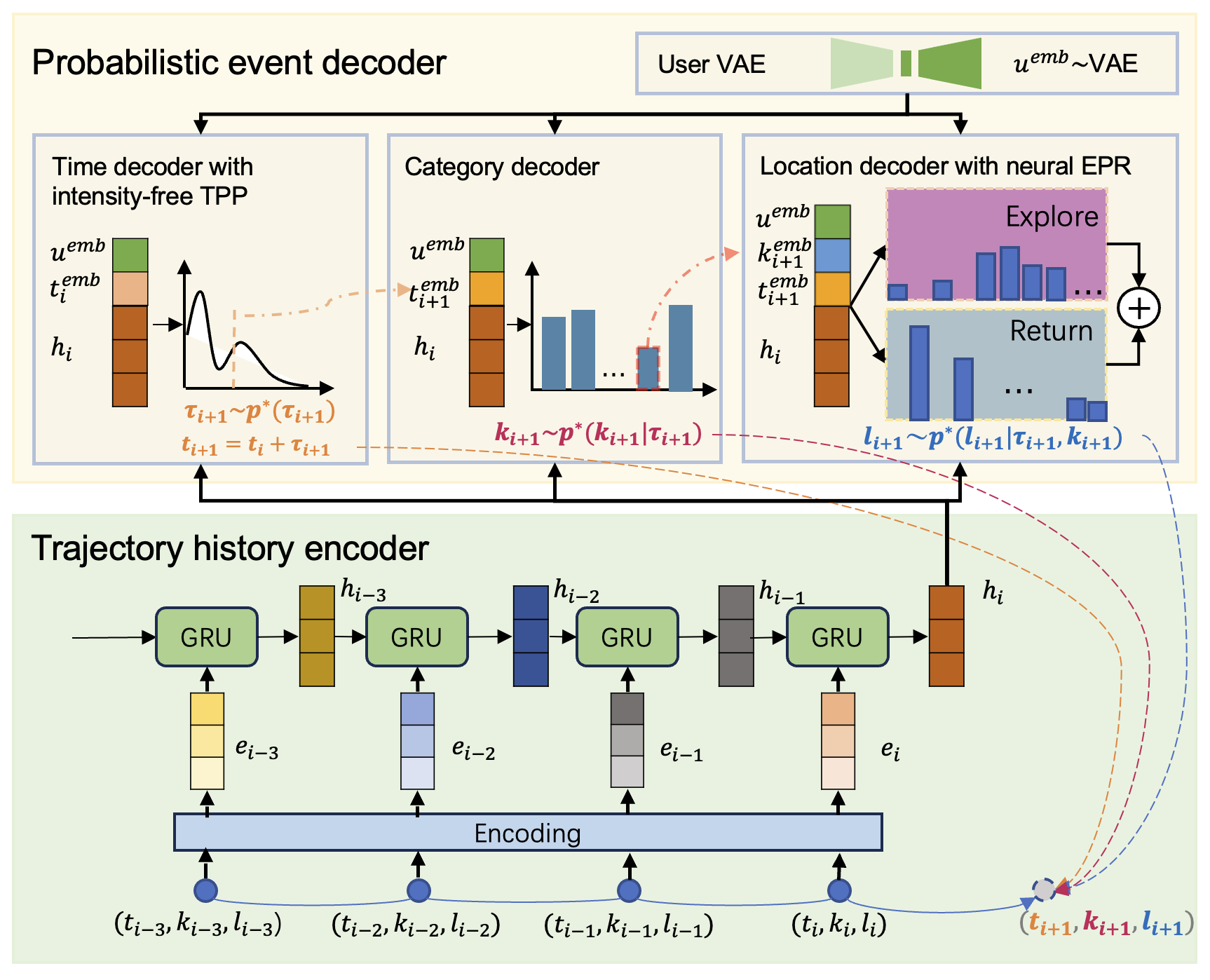}
	\vspace{-2em}
    \caption{An overview of our MIRAGE with 1) a trajectory history encoder and 2) a probabilistic event decoder integrating a time decoder with intensity-free TPP, a category decoder, a location decoder with neural EPR, and a user VAE.}
    \label{MIRAGE_archi}
    \vspace{-1em}
\end{figure}

\subsection{Overview}
Figure \ref{MIRAGE_archi} shows the overview of MIRAGE consisting of two components. To imitate the human decision-making process, a trajectory history encoder learns from the sequences of events using a recurrent neural network. Afterward, based on the output hidden states of the trajectory history encoder, a probabilistic event decoder generates the probability distributions for the next event time, activity categories, and locations in a cascading manner, where the latter distribution depends on the samples drawn from the former distribution. Notably, the distribution of the next event time is modeled using an intensity-free TPP, while the distribution of the next location is modeled by a neural EPR model. In addition, we also use a Variational Autoencoder to model the distribution of individual preferences on trajectory (User VAE).

% To generate one event of a human trajectory, we first sample an event time from the neural TPP based on the encoded trajectory history, and then sample the activity category further conditioned on the sampled event time. Afterward, we seamlessly integrate the EPR model parameterized by neural networks, mimicking the human decision-making process to choose either exploring unvisited locations or returning to previously visited locations, conditioned on the sampled time and activity categories.

 % where $x_i$ is a semantic spatiotemporal point defined as a tuple ($l_i$, $k_i$, $t_i$) consisting of a location  (POI) $l_i$, a timestamp $t_i$, and a semantic activity category $k_i$., 
 
% According to the definition, a trajectory can be considered as a realization of a marked TPP, and we utilized the neural temporal point process to model the distribution of check-in time and the activity semantics. This procedure involves 1) An event encoder which, for each event generates a fixed size embedding. 2) A history encoder, which generates a fixed size history embedding from past event representations;  3) A decoder, which parametrizes a log-normal mixture decoder parametrizes the conditional density of inter-arrival times as a mixture of log-normals using event embedding and history information, and activity semantics conditioned on generated time. For easy understanding, we introduce these components, respectively.

% \subsubsection{Event encoding}

\subsection{Trajectory History Encoder}
Representing a human trajectory as a sequence of events \{$x_1$, $x_2$, ..., $x_n$\}, the trajectory history encoder first encodes individual event $x_i = (t_i, k_i, l_i)$ to event embedding $e_i$, and then adopts a Gated Recurrent Unit (GRU) to encode the sequence of event embeddings. Specifically, for each event $(t_i, k_i, l_i)$, we first encode the timestamp $t_i$ by converting it to the log inter-event time $ \log(\tau_i) = \log (t_i - t_{i-1})$. This lossless conversion is to fit the formulation of the intensity-free TPP; however, the actual timestamp $t_i$ is still useful in the decoding process, and we will discuss this point later. We also encode the category $k_i$ and location $l_i$ using two embedding layers $E_k$ and $E_l$, respectively.
\begin{equation}
    e_{i}^{t} = log(\tau_i) , \quad
    e_{i}^{k} = E^k k_i , \quad
    e_{i}^{l} = E^l l_i 
\end{equation}
The event embedding $e_{i}$ is obtained by concatenating $e_{i}^{t}$, $e_{i}^{k}$ and $e_{i}^{l}$:
\begin{equation}
    e_{i} = [e_{i}^{t};e_{i}^{k};e_{i}^{l}]
\end{equation}
% \subsubsection{History encoding}
After obtaining a sequence of event embeddings, we then encode it using a recurrent neural network of GRU:
\begin{equation}
\label{encoded_history}
    h_{i} = g(h_{i-1},e_{i}) 
\end{equation}
where $g$ represents the recurrent updating function of GRU. The output hidden state $h_i$ encodes all the trajectory history until $t_i$.

% $e_{i}^{t}$ based on the inter-event time $\tau_i = (t_i - t_{i-1})$ and category and POI feature $e_{i}^{k}$ and $e_{i}^{l}$ encoded by embedding layers (learnable embedding matrices $E_k$ and $E_l$) based on the activity semantics $k_i$ and POI id $l_i$, respectively:

% A crucial feature of temporal point processes is that the time $\tau_i = (t_i - t_{i-1}, l_{i-1})$ until the next event may be influenced by all the events that happened before. To accurately estimate the distributions of the next inter-times and activity semantics of future events, we need to encode that event history (a variable-sized set) to a fixed-dimensional vector. In this context, we employ recurrent architecture (e.g. GRU), for example, after an event $(t_{i-1}, k_{i-1})$ occurs, the new history embedding $h_i$ is constructed by sequentially updating the history embedding $h_{i-1}$ with event embedding $e_{i-1}$:
% \begin{equation}
% \label{encoded_history}
%     h_{i} = g(h_{i-1},e_{i-1}) 
% \end{equation}
% Where $g$ refers to the update function of the GRU.

% \footnote{As the next event time $t_{i+1}=t_i+\tau_{t+1}$ can be computed in closed-form,} 
\subsection{Probabilistic Event Decoder}
Based on the output hidden state $h_i$, the probabilistic event decoder generates the probability distribution for the next inter-event time $p^*(\tau_{i+1})$, activity category $p^*(k_{i+1}|\tau_{i+1})$, and location $p^*(l_{i+1}|\tau_{i+1},k_{i+1})$ in a cascading manner, where the latter distribution depends on the samples drawn from the former distributions. Note that here (*) denotes the conditioning on the trajectory history, while the conditional probability emphasizes on conditioning on drawn samples.

% Notably, the distribution of the next event time is modeled using an intensity-free TPP, while the distribution of the next location is modeled by a neural EPR. In addition, we also use a Variational Autoencoder to model the distribution of individual preferences on trajectory (User VAE). 

\subsubsection{\textbf{Time decoder with intensity-free TPP}}
Following the intensity-free TPP \cite{shchur2019intensity}, we use a mixture of log-normal distributions to characterize the conditional probability density function of inter-event time $p(\tau)$. 
\begin{equation}
    p(\tau | \omega,\mu,\sigma) = \sum_{m=1}^{M} \omega_m \frac{1}{\tau \sigma_m \sqrt{2 \pi}} \exp \begin{pmatrix} -\frac{(\log \tau - \mu_m)^2}{2\sigma_{m}^{2}} \end{pmatrix} 
\end{equation}
where $M$ is the number of components of the mixture, $\omega_m$ denotes the weight of each component, $\mu_m$ and $\sigma_m$ are the logarithmic mean and standard deviation of each component.

Moreover, beyond the intensity-free TPP, we have two further design considerations to better accommodate the human trajectories. First, the original intensity-free TPP is defined on inter-event times only (as mentioned in the trajectory history encoder), which ignores the actual timestamp $t_i$ that could play a crucial role in human mobility modeling. For example, human trajectories exhibit diurnal rhythms, where few events are observed during the night on weekdays; in other words, if an event is observed at $t_i$ at midnight, the next event will probably be in the morning of the next day. This implies that $p^*(\tau_{i+1})$ should be skewed to the sleep duration here. Therefore, we transform the timestamp $t_i$ into an hour-in-week timestamp embedding $t_i^{emb}$ where $1 \leq i \leq 168$ (i.e., one of the 168 hours in a week), and $t_i^{emb}$ is used to further condition $p^*(\tau_{i+1})$. The hour-in-week granularity is selected to capture both daily and weekly dynamics as suggested by \cite{yang2019revisiting}. Second, human trajectories are intrinsically generated under individual preferences \cite{long2023practical}. For example, user check-ins at POIs on social networks have been widely used for user preference modeling \cite{ye2010location}. Therefore, we also define an individual preference embedding $u^{emb}$ for each trajectory to condition $p^*(\tau_{i+1})$.

% (we assign each trajectory a unique ID and embed the trajectory ID corresponding to individual preference)

Then, we concatenate the hidden state of trajectory history $h_i$, the hour-in-week timestamp embedding $t_i^{emb}$, and the individual preference embedding $u^{emb}$ as the overall time context for the conditional probability density function of inter-event time $p^*(\tau_{i+1})$:
\begin{equation}
        c_i^{\tau} = [h_{i};t_{i}^{emb};u^{emb}]
\end{equation}
We then compute the parameters of the distribution $p^*(\tau_{i+1})$ using an affine function of $c_i^{\tau}$:
\begin{equation}
    \begin{cases}
        \begin{aligned}
            &\omega_i = \mbox{softmax} (V_{\omega}c_i + b_{\omega}) \\
            &\mu_i = V_{\mu}c_i + b_{\mu} \\
            &\sigma_i = \exp(V_{\sigma}c_i + b_{\sigma})
        \end{aligned}
    \end{cases}
\end{equation}
where the softmax and exp transformations are used to enforce the constraints on the distribution parameters ( $\sum_{m}\omega_m = 1$, $\omega_m \geq 0$, and $\sigma_m \textgreater 0$), and $\{V_{\omega}, b_{\omega}, V_{\mu},b_{\mu}, V_{\sigma}, b_{\sigma}\}$ are learnable parameters. 

Finally, we sample a next inter-event time $\tau_{i+1}$ according to the log-normal mixture $p^*(\tau_{i+1}) = p(\tau_{i+1}|\omega_i, \mu_i, \sigma_i)$, and also obtain the corresponding next event time $t_{i+1} = t_{i} + \tau_{i+1}$.

\subsubsection{\textbf{Category decoder}}
After obtaining the next inter-event time $\tau_{i+1}$, we sample an activity category. In this step, we use a slightly different (category) context vector $c_i^{k}$ by replacing the timestamp embedding $t_{i}^{emb}$ in the time context vector $c_i^{\tau}$ with the embedding of the sampled next event time $t_{i+1}^{emb}$, as follows:
\begin{equation}
        c_i^{k} = [h_{i};t_{i+1}^{emb};u^{emb}]
\end{equation}
This is because the next activity category $k_{i+1}$ is highly correlated with its corresponding event time $t_{i+1}$; for example, the activity category in the noon time should probably be ``food'' (one of the nine categories on Foursquare \cite{yang2019revisiting}). Then, we compute a categorical distribution conditioned on the category context vector $c_i^{k}$:
\begin{equation}
    p^*(k_{i+1}|\tau_{i+1}) = \mbox{softmax} ( \mbox{MLP}_{\phi} (c_i^{k}) )
    \label{eq_cate_decoder}
\end{equation}
where $\mbox{MLP}_{\phi}$ refers to a multi-layer perception with parameters $\phi$ and softmax transforms its output to a categorical probability distribution. We finally sample a next event category $k_{i+1}$ from $p^*(k_{i+1}|\tau_{i+1})$. Note that as the next event time $t_{i+1}=t_i+\tau_{t+1}$ can be computed in closed-form from the sampled inter-event time $\tau_{t+1}$, we keep using $\tau_{i+1}$ as the condition of $p^*(k_{i+1}|\tau_{i+1})$ for emphasizing the sampling dependences.

\subsubsection{\textbf{Location decoder with neural EPR}}
After obtaining the next event time $t_{i+1}$ and category $k_{i+1}$, we sample the next location by designing a neural EPR model. Specifically, our location decoder follows a two-step design. First, we sample a binary decision on two competing modes: exploration (visiting a new location that is not in the trajectory history) or return (visiting a previously visited location in the trajectory history). Second, based on the sampled mode, we then sample a location from the corresponding candidate locations of the mode. We present the detailed design below.

\noindent \textbf{Exploration/Return mode sampling.}
In this step, we learn to sample one mode based on an extended context vector $c_{i}^{l}$, by further adding the embedding of the sampled next event category $k_{i+1}^{emb}$ to the context vector of category encoder $c_{i}^{k}$:
\begin{equation}
\label{next_l_context}
    c_{i}^{l} = [h_i;t_{i+1}^{emb};k_{i+1}^{emb};u^{emb}]\\  
\end{equation}
Subsequently, we feed this context vector to an MLP followed by a softmax function, to output a distribution over the two modes of $explore$ and $return$:
\begin{equation}
    p^* ( z |\tau_{i+1}, k_{i+1}) = \mbox{softmax} ( \mbox{MLP}_{\theta} (c_{i}^{l}))
\end{equation}
Finally, we sample one mode $z$ according to $p^* ( z |\tau_{i+1}, k_{i+1})$. Note that for the first event in a trajectory sequence, the exploration mode is selected, because of the empty trajectory history.

\noindent \textbf{Location sampling in the exploration mode.}
In the exploration mode $z=explore$, we compute a categorical distribution over all previously unvisited locations, which is obtained by feeding the location context vector $c_{i}^{l}$ to an MLP followed by a softmax function:
% \begin{align}
%     &L_e = MLP_{\xi}(c_{i}^{l})\\
%     &L_e[l] = - \inf, \mbox{where}\ l \in \{x_j | x_j \in X, j < i+1\} \\ \label{eq_selection_e}
%     &p^*_{l}(l|\tau_{i+1}, k_{i+1}, z=explore) = \mbox{softmax}(c_e)
% \end{align}
\begin{equation}
\begin{aligned}
   p^* (l_{i+1} |\tau_{i+1}, &k_{i+1}, explore)  = \mbox{softmax} ( \mbox{MLP}_{\xi}(c_{i}^{l})), \\ &\mbox{where}\ l \notin \{x_j | x_j \in X, j < i+1\}
\end{aligned}
\end{equation}

From this, we can sample a previously unvisited location $l$.
%  according to $p^*(l_{i+1} |\tau_{i+1}, k_{i+1}, explore)$

% Note that for the sake of notation consistency, $p^*_{l}$ represents the probability distribution over all locations. We thus set the MLP output values $L_e[l]$ that are associated with the visited locations in the trajectory $l \in \{x_j | x_j \in X, j < i+1\}$ to $-\inf$, corresponding to a zero probability in $p^*_{l}$ after softmax.

% Note that for the sake of notation consistency, $p^*_{l}$ represents the probability distribution over all locations. We thus set the MLP output values $L_e[l]$ that are associated with the visited locations in the trajectory $l \in \{x_j | x_j \in X, j < i+1\}$ to $-\inf$, corresponding to a zero probability in $p^*_{l}$ after softmax.

\noindent \textbf{Location sampling in the return mode.} 
In the return mode $z=return$, we consider the temporal periodicity of human behaviors, where individuals tend to revisit previously visited locations under a regular temporal distance, such as returning home daily (with a temporal distance of 24 hours) \cite{song2010modelling}. Therefore, the return mode is designed to imitate such revisiting behaviors. Specifically, instead of imposing specific constraints or predefined temporal periodicity as \cite{yang2020location1, deng2023robust}, we use temporal distance embeddings to learn the probability distribution of returning to previously visited locations. We define the temporal distance under an hour granularity between the $(i+1)$-th and $j$-th events as $\Delta^t_{i+1,j} = |t_{i+1} - t_j|  \in \mathbb Z_{\ge 0}$. Subsequently, for the next event time $t_{i+1}$, we obtain its temporal distances to all previous events as follows:
\begin{equation}
    \Delta^t = [\Delta^t_{i+1,1}, \Delta^t_{i+1,2}, ... , \Delta^t_{i+1,i}]
\end{equation}
We then define a learnable embedding for each hourly time distance, which is fed to an MLP followed by softmax to output the categorical distribution of returning to the locations of the previous events:
\begin{equation}
\begin{aligned}
    p^*(l_{i+1}|\tau_{i+1}, &return) = \mbox{softmax} (\mbox{MLP}_\eta([\Delta^{emb}_{i+1,1}, \Delta^{emb}_{i+1,2}, ... , \Delta^{emb}_{i+1,i}])), \\
    &\mbox{where}\ l \in \{x_j | x_j \in X, j < i+1\}
\end{aligned}
\end{equation}
Note that here $p^*(l_{i+1}|\tau_{i+1}, return)$ is not conditioned on the sampled next event category $k_{i+1}$ because the revisiting behavior depends only on the next event time. We then sample a previously visited location $l$ according to $p^*(l_{i+1}|\tau_{i+1}, return)$. Our experiments also reveal that this learnt probability indeed exhibits a clear daily return pattern over different time distances (see Figure \ref{weight_dist}).
% (see Appendix \ref{sec:weight_dist})

In summary, our location decoder with neural EPR samples a next location $l_{i+1}$ according to:
\begin{equation}
\begin{aligned}
p^*(l_{i+1} &| \tau_{i+1},k_{i+1}) = \\
& p^* (explore |\tau_{i+1}, k_{i+1}) \cdot p^* (l_{i+1} |\tau_{i+1}, k_{i+1}, explore) \\ 
&+ p^* ( return |\tau_{i+1}, k_{i+1})  \cdot p^*(l_{i+1}|\tau_{i+1}, return)
\end{aligned}
\end{equation}

% Specifically, we compute the temporal distance matrix $\Delta^t \in \mathbb{R}^{1 \times (i-1)}$ between all of the historical check-in times $t_j$ (for $j < i$) and $t_i$ (the time of the upcoming check-in):
% \begin{equation}
%     \Delta^t = [\Delta^t_{i,1}, \Delta^t_{i,2}, ... , \Delta^t_{i,i-1}]
% \end{equation}
% Based on the computed temporal distance matrix, for each time interval, we calculate its corresponding return intensity using a multi-layer perceptron (MLP). Then, we use a softmax function to normalize return intensity into a probability distribution. After obtaining the probability distribution, we sample from the previously visited locations corresponding to each time interval based on this distribution to generate the returned location.

\subsubsection{User VAE}
Additionally, we also train a Variational Autoencoder (VAE) to learn to generate novel individual preference embeddings on trajectories, to avoid exposing the learnt user preference embeddings from the real dataset in the generation process. In the sampling stage, the trained user VAE generates individual preference embedding from random noises.

\subsection{Model Training and Trajectory Generation}
\subsubsection{Model Training} 
The training objective of MIRAGE is to minimize the Negative Log Likelihood (NLL) of all sequences of events in a human trajectory dataset $\mathcal{D}$.  
\begin{equation}
    \mathcal{L} (\mathcal{D}) = -\sum_{X \in \mathcal{D}} \sum_{x_i \in X} \left( \ln p^*(\tau_i) + \ln p^*(k_{i}|\tau_{i}) + \ln p^*(l_{i}|\tau_{i},k_{i}) \right)
\end{equation}
In the training process, the above conditional probability distribution is conditioned on the real data without sampling. After training this objective function, we then train the user VAE on the learnt individual preference embeddings. The complexity of our model is discussed in Appendix \ref{complexity_appendix}.

%The initial hidden state $h_0$ of GRU is a learnable embedding. 

\subsubsection{Trajectory Generation}
Our trajectory generation process is conducted via a series of sampling steps without any real data as input. Specifically, the user VAE first generates an individual preference embedding. Afterward, the trajectory is generated iteratively based on the encoded trajectory history as follows: 1) the time decoder samples an inter-event time $\tau_i$ according to $p^*(\tau_i)$; 2) the category decoder samples an activity category $k_{i}$ conditioned on $\tau_i$ according to $p^*(k_{i}|\tau_{i})$; and 3) the location decoder samples a location $l_{i}$ conditioned on $\tau_i$ and $k_{i}$ according to $p^*(l_{i}|\tau_{i},k_{i})$. Note that the initial hidden state $h_0$ is learnt during training, and the generation process terminates until a desired time length.

\section{Experiments}

\subsection{Experimental Settings}
\subsubsection{Dataset}
We conduct extensive experiments on three user trajectory datasets collected from a location-based social network Foursquare \cite{yang2019revisiting,yang2020lbsn2vec++}, in three respective cities Tokyo (\textbf{TKY}), Istanbul (\textbf{IST}), and New York City (\textbf{NYC}). Table \ref{dataset} shows the dataset statistics.

%Each dataset includes anonymous user IDs, timestamps, categories, and POIs. For the category information of POIs, we have specifically selected nine main categories: Arts \& Entertainment, College \& University, Event, Food, Nightlife Spot, Outdoors \& Recreation, Professional \& Other Places, Shop \& Service, and Travel \& Transport.

\begin{table}[]
\centering
\caption{Dataset statistics} \label{dataset}
\vspace{-1em}
\small
\begin{tabular}{l|ccc}
\hline
& \textbf{TKY} & \textbf{IST} & \textbf{NYC} \\
\hline
\#Sequences & 8,890 & 14,380 & 17,682 \\
%\hline
\#POIs & 7,590 & 16,809 & 15,723 \\
%\hline
\#Categories & 9 & 9 & 9 \\
\hline
\end{tabular}
\vspace{-1em}
\end{table}

\subsubsection{Baselines} \label{sec:baselines}
We consider the following state-of-the-art baselines of three categories: statistical models \textbf{Semi-Markov} \cite{korolyuk1975semi} and
\textbf{TimeGeo} \cite{jiang2016timegeo}; neural TPP \textbf{RMTPP} \cite{du2016recurrent}, \textbf{ERTPP} \cite{ertpp}, \textbf{THP} \cite{zuo2020transformer}, \textbf{ActSTD} \cite{yuan2022activity} and \textbf{LogNormMix} \cite{shchur2019intensity}; deep learning generative models \textbf{LSTM} \cite{huang2015bidirectional}, \textbf{SeqGAN} \cite{yu2017seqgan}, \textbf{MoveSim} \cite{feng2020learning}, \textbf{VOLUNTEER} \cite{long2023practical} and \textbf{DiffTraj} \cite{zhu2024difftraj}. The baseline details are in Appendix \ref{sec:app_baseline}.

\subsubsection{Statistical and Distributional Similarity Metrics}
We adopt five popular metrics \cite{feng2020learning,ouyang2018non} to evaluate the resemblance between real and generated trajectories in different aspects. \textbf{Distance} measures the distance between successive locations in a trajectory. \textbf{Radius} of gyration is calculated as the root mean squared distance of all locations from the central one in a trajectory. \textbf{Interval} is computed as time intervals between successive events in a trajectory. \textbf{DailyLoc} computes the unique locations visited by users. \textbf{Category} computes the overall distribution of the POI categories. We use the Jensen-Shannon divergence (JSD) \cite{fuglede2004jensen} as the similarity metric between the distributions of real and generated trajectories.
% \textbf{G-Rank} computes the visiting frequency of the top 100 locations among all users' trajectories.
% \begin{equation}
% JSD(P \| Q) = \frac{1}{2} \left(D_{KL}(P \| M) + D_{KL}(Q \| M)\right)  
% \end{equation}
% where $D_{KL}(\cdot \| \cdot)$ is the Kullback-Leibler divergence and $M$ is the average distribution defined as $M = \frac{1}{2}(P + Q)$.

% of spatial, temporal, and user distributions

\begin{table*}[]
\caption{Performance on Statistical and Distributional Similarity Metrics}
\label{statistical_performance}
\vspace{-1em}
\setlength{\tabcolsep}{0.20em}
\small
\begin{tabular}{l|ccccc|ccccc|ccccc}
\hline
& \multicolumn{5}{c|}{TKY} & \multicolumn{5}{c|}{IST} & \multicolumn{5}{c}{NYC} \\
\hline
Method & Distance & Radius & Interval & DailyLoc & Category & Distance & Radius & Interval & DailyLoc & Category & Distance & Radius & Interval & DailyLoc & Category \\
\hline
Semi-Markov & 0.6931 & 0.5840 & 0.1196 & 0.6931 & 0.0791 & 0.6931 & 0.5545 & 0.2041 & 0.6931 & 0.0295 & 0.6931 & 0.5097 & 0.2072 & 0.6930 & 0.0135 \\
Time Geo & 0.5307 & 0.5712 & 0.0467 & 0.4441 & 0.0140 & 0.5984 & 0.6339 & 0.0345 & 0.3644 & 0.0124 & 0.4437 & 0.5673 & 0.0387 & 0.1982 & 0.0144 \\
\hline
RMTPP & 0.1988 & 0.1724 & 0.4166 & 0.5778 & 0.0113 & 0.1864 & 0.2008 & 0.4151 & 0.5650 & 0.0322 & 0.1745 & 0.2288 & 0.4422 & 0.3728 & 0.0141 \\
ERTPP & 0.6807 & 0.4021 & 0.1249 & 0.6636 & 0.0187 & 0.6699 & 0.3601 & 0.2110 & 0.6762 & 0.0236 & 0.6783 & 0.4395 & 0.2127 & 0.6824 & 0.0169 \\
THP & 0.3604 & 0.0858 & 0.2290 & 0.6183 & 0.0815 & 0.3476 & 0.4975 & 0.1122 & 0.2414 & 0.0340 & 0.4239 & 0.5109 & 0.0738 & 0.6235 & 0.0544 \\
ActSTD &0.2470 &0.1921 &0.0272 &0.1931 &0.0231 &0.1710 &0.1225 &0.0098 &0.1325 &0.0231 &0.1884 &0.1861 &0.0152 &0.1360 &0.0131 \\
LogNormMix & 0.3898 & 0.2585 & 0.0186 & 0.2114 & 0.0056 & 0.2567 & 0.1787 & 0.0299 & 0.1547 & 0.0152 & 0.2172 & 0.1751 & 0.0352 & 0.1120 & 0.0205 \\
\hline
LSTM & 0.2745 & 0.1983 & 0.0919 & 0.1581 & 0.0143 & 0.3640 & 0.2516 & 0.0758 & 0.1960 & 0.0671 & 0.2876 & 0.2425 & 0.0476 & 0.2501 & 0.0133 \\
SeqGAN & 0.3006 & 0.2616 & 0.0507 & 0.2406 & 0.0326 & 0.1548 & 0.1734 & 0.0584 & 0.1279 & 0.0125 & 0.3023 & 0.2897 & 0.1206 & 0.3102 & 0.0078 \\
MoveSim & 0.3623 & 0.3018 & 0.0619 & 0.2761 & 0.0136 & 0.4549 & 0.3176 & 0.1537 & 0.4076 & 0.0527 & 0.3454 & 0.3700 & 0.0967 & 0.2293 & 0.0527 \\
VOLUNTEER & 0.5098 & 0.3208 & 0.0273 & 0.2630 & 0.0184 & 0.2498 & 0.2623 & 0.0415 & 0.1031 & 0.0171 & 0.3167 & 0.2845 & 0.0443 & 0.1456 & 0.0201 \\
DiffTraj & 0.2209 & 0.2653 & 0.1388 & 0.2684 & 0.0769 & 0.1974 & 0.0713 & 0.2137 & 0.1513 & 0.0237 & 0.1589 & 0.1039 & 0.1949 & 0.2361 & 0.0074 \\
\hline
MIRAGE & \textbf{0.1295} & \textbf{0.0330} & \textbf{0.0037} & \textbf{0.0229} & \textbf{0.0038} & \textbf{0.0714} & \textbf{0.0375} & \textbf{0.0033} & \textbf{0.0100} & \textbf{0.0045} & \textbf{0.0622} & \textbf{0.0415} & \textbf{0.0050} & \textbf{0.0185} & \textbf{0.0024} \\
\hline
\end{tabular}
\vspace{-0.5em}
\end{table*}

\begin{table*}
\caption{Performance in the Task-Based Evaluation on MAPE}
\label{down_performance_L1}

\vspace{-1em}
\small
\begin{tabular}{l|cccc|cccc|cccc}
\hline
& \multicolumn{4}{c|}{TKY} & \multicolumn{4}{c|}{IST} & \multicolumn{4}{c}{NYC} \\
\hline
Method & LocRec & NexLoc & SemLoc & EpiSim & LocRec & NexLoc & SemLoc & EpiSim & LocRec & NexLoc & SemLoc & EpiSim \\
\hline
Semi-Markov & 0.8335 & 0.9919 & 0.4784 & 0.6406 & 0.9165 & 0.9969 & 0.5179 & 0.5652 & 0.8410 & 0.9951 & 0.4472 & 2.5354 \\
Time Geo & 0.7487 & 1.7867 & 0.2725 & 0.8531 & 3.4468 & 3.4048 & 0.3962 & 0.8838 & 1.0636 & 3.0667 & 0.2691 & 0.8122 \\
\hline
RMTPP & 0.8694 & 0.8616 & 0.2888 & 0.9402 & 0.7237 & 0.8180 & 0.4412 & 0.8599 & 0.8879 & 0.9406 & 0.3786 & 0.8801 \\
ERTPP & 0.4367 & 0.8181 & 0.1878 & 0.5592 & 0.3605 & 0.7854 & 0.3596 & 0.7666 & 0.6810 & 0.9156 & 0.5158 & 3.4278 \\
THP & 1.4991 & 2.2344 & 0.3985 & 0.3496 & 0.5154 & 0.8491 & 0.4963 & 0.9836 & 3.2324 & 3.7909 & 0.3496 & \textbf{0.6705} \\
ActSTD &0.5341 &0.6174 &0.3738 &0.2366 &0.4632 &0.5734 &0.4454 &0.4095 &0.6648 &0.5953 &0.4158 &0.8923 \\
LogNormMix & 0.4549 & 0.7148 & 0.1822 & 0.1480 & 0.4374 & 0.8592 & 0.3862 & \textbf{0.1620} & 0.7031 & 0.8702 & 0.4146 & 0.7216 \\
\hline
LSTM & 0.4816 & 0.7230 & 0.1631 & 0.1490 & 0.4025 & 0.6398 & 0.4016 & 0.2745 & 0.6884 & 0.5246 & 0.2889 & 0.6809 \\
SeqGAN & 0.4460 & 0.4514 & 0.2909 & 0.2041 & 0.8808 & 0.6971 & 0.4172 & 0.2216 & 1.0619 & 0.5271 & 0.5199 & 2.3157 \\
MoveSim & 0.7371 & 0.3347 & 0.2791 & 0.1790 & 5.0527 & 0.5815 & 0.4978 & 0.5957 & 1.1737 & 0.7824 & 0.3368 & 2.2911 \\
VOLUNTEER & 0.7277 & 0.5392 & 0.3050 & 0.4577 & 0.4796 & 0.5417 & 0.3750 & 0.2299 & 0.6695 & 0.8459 & 0.4831 & 1.6571 \\
DiffTraj & 0.7649 & 1.0815 & 0.4555 & 0.6752 & 0.7345 & 1.1090 & 0.4840 & 0.8263 & 0.6622 & 1.9322 & 0.4059 & 0.7631 \\
\hline
MIRAGE & \textbf{0.3485} & \textbf{0.3173} & \textbf{0.1433} & \textbf{0.1390} & \textbf{0.2947} & \textbf{0.3046} & \textbf{0.2198} & 0.2170 & \textbf{0.4855} & \textbf{0.2299} & \textbf{0.0961} & 0.7609 \\
\hline
\end{tabular}
\vspace{-0.5em}
\end{table*}

\subsection{Task-Based Evaluation Protocol}
We introduce our proposed task-based evaluation protocol to comprehensively benchmark trajectory generative models. Specifically, the ultimate utility of human trajectory generation is to support different downstream tasks in practice; subsequently, the benchmark objective is to evaluate whether the generated trajectories are similar to real trajectories when being used to conduct different downstream tasks. In the current literature, existing works all use heuristically designed downstream tasks with one specific technique to solve a task \cite{long2023practical, yuan2022activity, feng2020learning, jiang2023continuous, cao2021generating}, and thus lack a comprehensive view of utility benchmarks. In particular, heuristically designed downstream tasks may lead to unknown biases in the utility evaluation, as the performance of different techniques solving the same task often varies (as evidenced by our experiments in Appendix \ref{sec:app_task_bias}), and heuristically regarding the results of one technique as the benchmark is thus untrustworthy. 

Our proposed evaluation protocol implements four typical tasks: location recommendation, next location prediction, semantic location labeling, and epidemic simulation, which model user trajectory data in four aspects, user preferences on locations, sequential mobility patterns, collective traffic patterns, and spatiotemporal contact patterns, respectively. For each task, we choose multiple state-of-the-art techniques to conduct experiments and report the results on multiple metrics, to average out the biases of individual techniques and metrics. We then measure the paired performance discrepancy between the real and generated trajectories using Mean Absolute Percentage Error (\textbf{MAPE}) and Mean Squared Percentage Error (\textbf{MSPE}), which serve as final benchmarks to assess the ultimate utility of the generated trajectories. In the following, we first present our dataset settings, followed by the details of each task.

%We present the details of individual tasks below.

\subsubsection{Dataset Settings}
Unlike some previous works \cite{long2023practical,yuan2022activity,jiang2023continuous} that use the generated trajectories to augment the real trajectories (combining generated and real data in certain proportions) and then perform the downstream tasks, we put one step forward to directly perform the tasks on the generated trajectories only, without exposing any real trajectories. Our dataset setting is more strict which completely avoids the leakage of real human trajectories. In our experiments, for each real trajectory dataset, we generate a synthetic dataset having the same number of trajectories as the real dataset using each trajectory generative model. We then perform downstream tasks on both real and generated datasets separately, to evaluate whether the two datasets encode the same amount of information for supporting different downstream tasks.

% To maintain the applicability of synthetic data in real-world tasks and applications, the performance of downstream tasks of the synthetic data must be closely similar to the real data. Previous works \cite{long2023practical,yuan2022activity,feng2020learning,jiang2023continuous,cao2021generating} usually employ a single model for downstream tasks, potentially introducing bias, our proposed evaluation protocol includes multiple tasks and conducts experiments on each dataset, offering a more comprehensive evaluation. Our downstream tasks can be divided into three categories: location recommendation tasks, sequence recommendation tasks, and infectious disease simulation experiments. These tasks hold significant importance in the field of human mobility and for clarity, we introduce the details of three downstream tasks, respectively.
% or conduct the downstream task on a single dataset,

\subsubsection{Location Recommendation Task} (\textbf{LocRec}) suggests new (previously unvisited) locations for users by modeling \textit{user preferences on locations} from trajectory data \cite{ye2010location}. Following the default setting of \cite{recbole}, we transform a trajectory dataset into a set of (user, location, visit\_count) triplets and then split them into training/valid/test datasets under a ratio of 8:1:1. To discount the impact of the specific techniques and metrics, we consider five popular recommendation algorithms, i.e., BPR \cite{bpr}, DMF \cite{dmf}, LightGCN \cite{he2020lightgcn}, MultiVAE \cite{multivae}, and NeuMF \cite{neumf} (details in Appendix \ref{detail_locrec}), and report their performance on Mean Reciprocal Rank@N (MRR@N), Normalized Discounted Cumulative Gain@N (NDCG@N), hit@N (where N = 5 and 10). Finally, we compare the paired performance discrepancy between the real and generated trajectories (i.e., 30 paired results from five algorithms and six metrics each) using MAPE and MSPE.

\subsubsection{Next Location Prediction Task} (\textbf{NexLoc}) forecasts a user’s next location in the future by learning the \textit{sequential mobility patterns} from historical user trajectories \cite{noulas2012mining}.  Specifically, for one dataset, we chronologically split each trajectory into training/valid/test trajectories under a ratio of 8:1:1. We also consider five sequence prediction algorithms, i.e., FPMC \cite{fpmc}, BERT4Rec \cite{sun2019bert4rec}, Caser \cite{caser}, SRGNN \cite{srgnn}, and SASRec \cite{sasrec} (details in Appendix \ref{detail_seqrec}), and report their performance on MRR@N, NDCG@N, hit@N (where N = 5 and 10). Finally, we compare the paired performance discrepancy between the real and generated trajectories using MAPE and MSPE.

% This task evaluates whether the generated trajectory can preserve a similar amount of \textit{information on sequential mobility patterns} as the real trajectories.

% We conducted experiments on five representative sequential recommendation methods: \textbf{FPMC} \cite{fpmc}, \textbf{BERT4Rec} \cite{sun2019bert4rec}, \textbf{Caser} \cite{caser}, \textbf{SRGNN} \cite{srgnn}, \textbf{SASRec} \cite{sasrec}. The details of these algorithms can be found in the Appendix \ref{detail_seqrec}.

% In this task, we also employ RecBole as our evaluation tool to get the sequential recommendation performance of the real trajectory dataset and the synthetic trajectory dataset, respectively. After obtaining the performance of both the real trajectory dataset and the synthetic trajectory dataset on MRR@N, NDCG@N, and Hit@N (where N = 5, 10). Then, we calculate the Mean Absolute Percentage Error (MAPE) and Mean Squared Percentage Error (MSPE) for all these metrics for their performance comparison between the synthetic dataset and the real dataset as our metric.

\begin{table*}
\caption{Ablation Study on Statistical and Distributional Similarity Metrics}
\label{res_ablation_study_statistical}
\vspace{-1em}
\small
\setlength{\tabcolsep}{0.20em}
\begin{tabular}{l|ccccc|ccccc|ccccc}
\hline
& \multicolumn{5}{c|}{TKY} & \multicolumn{5}{c|}{IST} & \multicolumn{5}{c}{NYC} \\
\hline
Method & Distance & Radius & Interval & DailyLoc & Category & Distance & Radius & Interval & DailyLoc & Category & Distance & Radius & Interval & DailyLoc & Category \\
MIRAGE-noTPP & 0.1673 & 0.0448 & 0.1436 & 0.0465 & 0.0043 & 0.1550 & 0.1449 & 0.2742 & 0.0551 & 0.0080 & 0.2275 & 0.0987 & 0.2453 & 0.1817 & 0.0105 \\
MIRAGE-noEPR & 0.1468 & 0.0752 & 0.0183 & 0.1276 & 0.0108 & 0.0720 & 0.0429 & 0.0336 & 0.0221 & 0.0063 & 0.0920 & 0.0565 & 0.0374 & 0.0955 & 0.0044 \\
MIRAGE-noIMI &0.1368 &0.0334 &0.0051 &0.0238 &0.0101 &0.0749 &\textbf{0.0356} &0.0047 &0.0342 &0.0084 &0.0660 &0.0416 &0.0081 &0.0194 &0.0096 \\
MIRAGE-TD &0.1339 &0.0429 &\textbf{0.0037} &0.0289 &0.0048 &0.0933 &0.0368 &0.0036 &0.0149 &0.0059 &0.0692 &\textbf{0.0413} &0.0064 &0.0206 &0.0033 \\
\hline
MIRAGE & \textbf{0.1295} & \textbf{0.0330} & \textbf{0.0037} & \textbf{0.0229} & \textbf{0.0038} & \textbf{0.0714} & 0.0375 & \textbf{0.0033} & \textbf{0.0100} & \textbf{0.0045} & \textbf{0.0622} & 0.0415 & \textbf{0.0050} & \textbf{0.0185} & \textbf{0.0024} \\
\hline
\end{tabular}
\vspace{-0.5em}
\end{table*}

\begin{table*}[]
\caption{Ablation study in the Task-Based Evaluation on MAPE}
\label{res_ablation_study_downstream}
\vspace{-1em}
\small
\begin{tabular}{l|cccc|cccc|cccc}
\hline
& \multicolumn{4}{c|}{TKY} & \multicolumn{4}{c|}{IST} & \multicolumn{4}{c}{NYC} \\
\hline
Method & LocRec & NexLoc & SemLoc & EpiSim & LocRec & NexLoc & SemLoc & EpiSim & LocRec & NexLoc & SemLoc & EpiSim \\ \hline
MIRAGE-noTPP & 0.3688 & 0.3293 & 0.1600 & 0.5806 & 0.4458 & 0.4380 & 0.2323 & 0.3768 & 0.6654 & 0.5663 & 0.1153 & 1.4793 \\
MIRAGE-noEPR & 0.3725 & 0.4025 & 0.1483 & 0.1409 & 0.3620 & 0.5077 & 0.2427 & 0.2342 & 0.6243 & 0.5243 & 0.1045 & 0.9413 \\
MIRAGE-noIMI &\textbf{0.3479} &0.3458 &0.1820 &0.1457 &0.5026 &0.3399 &0.3283 &\textbf{0.2148} &0.5437 &0.2807 &0.2808 &0.8046 \\
MIRAGE-TD &0.3557 &0.3732 &0.2336 &0.1574 &0.3526 &0.3625 &0.3232 &0.2490 &0.5003 &0.2381 &0.2517 &0.8387 \\
\hline
MIRAGE & 0.3485 & \textbf{0.3173} & \textbf{0.1433} & \textbf{0.1390} & \textbf{0.2947} & \textbf{0.3046} & \textbf{0.2198} & 0.2170 & \textbf{0.4855} & \textbf{0.2299} & \textbf{0.0961} & \textbf{0.7609} \\
\hline
\end{tabular}
\vspace{-0.5em}

\end{table*}

\subsubsection{Semantic Location Labeling Task} (\textbf{SemLoc}) assigns a semantic label (i.e., activity category) to a location based on the \textit{collective traffic pattern} of the location, extracted from users' trajectory data \cite{yang2016poisketch}. Specifically, for each location, we extract its weekly temporal traffic pattern with an hour granularity, resulting in a feature vector of size 168 where each entry represents the empirical probability of all users' visits to this location. For each trajectory dataset, we then split all POIs into training/valid/test datasets under a ratio of 8:1:1. As a classification problem in nature, we consider five typical classification algorithms, i.e., Decision Tree, Naive Bayes, K-Nearest Neighbors, Logistic Regression, and Support Vector Machine, and report their performance on Accuracy, F1-Micro, and F1-Macro scores. Finally, we compare the performance discrepancy between the real and generated trajectories using MAPE and MSPE.
% tree.DecisionTreeClassifier	naive_bayes.GaussianNB	neighbors.KNeighborsClassifier	svm.LinearSVC	linear_model.LogisticRegression

\subsubsection{Epidemic Simulation Task} (\textbf{EpiSim}) simulates the epidemic spreading over a contact network characterizing the \textit{spatiotemporal contact patterns} of user trajectories \cite{verma2021spatiotemporal}. The contact network is extracted from a trajectory dataset as a dynamic graph of users, and a directed edge from a user $p$ to a user $q$ represents that $p$'s visit precedes $q$'s visit to the same location during a day, indicating a potential chance of epidemic spreading from $p$ to $q$. Following the setting of recent works \cite{feng2020learning, yuan2022activity, lai2020effect}, we adopt the Susceptible–Exposed–Infected–Recovered (SEIR) model. We consider the simulation of COVID-19 using the parameters suggested by \cite{feng2020learning,yuan2022activity} and influenza using the parameters suggested by \cite{brauer2019models} (details in Appendix \ref{sec:app_episim}). We randomly select 50 individuals as exposed individuals, simulate the spread of the epidemic, and report the daily counts of Exposed, Infectious, and Recovered individuals as three metrics. We report the average results of 10 repeated simulations to discount the impact of the random selection of initially exposed individuals. Finally, we compare the performance discrepancy between the real and generated trajectories using MAPE and MSPE.

\subsection{Statistical \& Distributional Similarity}
\label{sec:exp_datasaurus}
% \subsection{Performance on Statistical \& Distributional Similarity Metrics}
Table \ref{statistical_performance} shows the performance comparison of MIRAGE and baselines on the three datasets. We observe that MIRAGE consistently achieves the best performance with the lowest JSD, yielding 59.0\%, 64.4\%, and 67.7\% improvement (reduction on JSD) over the best-performing baselines on TKY, IST, and NYC datasets, respectively. However, even though we selected four similarity metrics covering three different aspects of spatial (Distance and Radius), temporal (Interval), and user (DailyLoc) distributions, these metrics cannot fully reflect the ultimate utility of generated trajectories in supporting downstream tasks. For example, the consistent superiority of MIRAGE over baselines on these similarity metrics is still biased, because MIRAGE indeed underperforms some baselines in a few cases in our task-based evaluation, as we discuss below.

% Although these metrics have been widely used as benchmarks,

% Compared to the best-performing baselines, our MIRAGE achieves 48.2\%, 74.4\% and 67.4\% improvement on TKY ,IST and NYC datasets, respectively. We observe that Semi-Markov and TimeGeo, which require parameter estimation from the training data, exhibit worse performance. The LSTM performs well in the Interval metric due to its use of a time module that optimizes time predictions. However, MoveSim does not fare well, as it requires discrete-time input, potentially leading to information loss from the original data. LogNormMix demonstrates the second-best performance in most cases by directly modeling the distribution of both time and POI, however, it does not consistently outperform other baselines in downstream task metrics. VOLUNTEER performs well in the Interval metric, leveraging temporal point processes to model time distribution. As for metrics ranking 2nd, MIRAGE also delivers comparable performance compared to the best baseline.

\vspace{-0.5em}
\subsection{Task-Based Evaluation Performance}
Table \ref{down_performance_L1} shows the performance in our proposed task-based evaluation on MAPE (similar results on MSPE, shown in Appendix \ref{sec:addtional_result}). Each entry in this Table is the MAPE averaging over all metrics of all algorithms solving a task on a dataset, to average out the biases of individual techniques and metrics. For example, the MAPE of MIRAGE in the LocRec task on the TKY dataset is 0.3485, which is computed from 30 paired (real and generated) results from five algorithms and six metrics used in the LocRec task. 

We observe that MIRAGE achieves the best performance with the smallest MAPE in most cases. In general, compared to the best-performing baselines, our MIRAGE achieves 10.9\%, 16.7\%, and 33.4\% improvement on TKY, IST, and NYC datasets, respectively. In addition, we also see that in the EpiSim task, MIRAGE achieves second and even fourth places, on IST and NYC datasets, respectively, which departs from the the consistent superiority of MIRAGE over all baselines on the statistical and distributional similarity metrics.

\vspace{-0.5em}
\subsection{Ablation Study}
\label{ablation_study_sec}
% We conduct an ablation study on our proposed method MIRAGE, considering the following two variants. \textbf{MIRAGE-noTPP} is a variant of MIRAGE without the neural TPP, where we define a regression task to predict the next inter-event from the time context vector $c_i^{\tau}$. \textbf{MIRAGE-noEPR} is a variant of MIRAGE without neural EPR, where we directly sample a location from a categorical distribution conditioned on the location context vector $c_i^{l}$. \textbf{MIRAGE-noIMI} is a variant of MIRAGE uses $t_{i}^{emb}$ and $k_{i}^{emb}$ for sampling the next category and next POI. \textbf{MIRAGE-TD} uses temporal distance in the exploration mode. Tables \ref{res_ablation_study_statistical} and \ref{res_ablation_study_downstream} show the results on similarities and task-based evaluation on MAPE, respectively (similar results on MSPE in Appendix \ref{sec:addtional_result}). 

We conduct an ablation study on our proposed method MIRAGE, considering the following four variants. \textbf{MIRAGE-noTPP} is a variant of MIRAGE without the neural TPP, where we define a regression task to predict the next inter-event from the time context vector $c_i^{\tau}$. \textbf{MIRAGE-noEPR} is a variant of MIRAGE without neural EPR, where we directly sample a location from a categorical distribution conditioned on the location context vector $c_i^{l}$. \textbf{MIRAGE-noIMI} is a variant of MIRAGE uses $t_{i}^{emb}$ and $k_{i}^{emb}$ (rather than $t_{i+1}^{emb}$ and $k_{i+1}^{emb}$) for sampling the next category and next POI. \textbf{MIRAGE-TD} is a variant of MIRAGE that uses temporal distances (rather than timestamps) in the exploration mode. Tables \ref{res_ablation_study_statistical} and \ref{res_ablation_study_downstream} show the results on similarities and task-based evaluation on MAPE, respectively (similar results on MSPE in Appendix \ref{sec:addtional_result}).

We observe that MIRAGE consistently outperforms MIRAGE-noTPP on both similarities and task-based evaluation by 63.8\% and 30.0\% (on average over tasks and datasets), respectively, showing the effectiveness of neural TPPs modeling the event stochasticity of human trajectories. Second, MIRAGE outperforms MIRAGE-noEPR in most cases, with an improvement of 50.3\% and 17.8\% on similarities and task-based evaluation, respectively, which verifies the usefulness of our neural EPR model. Third, MIRAGE outperforms MIRAGE-noIMI in general, by 24.8\% and 18.1\% on similarities and task-based evaluation, respectively, since the next location should depend on the next category and time, rather than the current ones. Finally, MIRAGE outperforms MIRAGE-TD in most cases, by 15.2\% and 18.5\% on similarities and task-based evaluation, respectively. Because timestamps offer more direct information for location sampling in the exploration mode compared to temporal distances. Our ablation studies systematically validate our key design choices.

\begin{figure}
\centering
\vspace{-0.5em}
\subfigure[Real] {
\includegraphics[width=0.3\columnwidth]{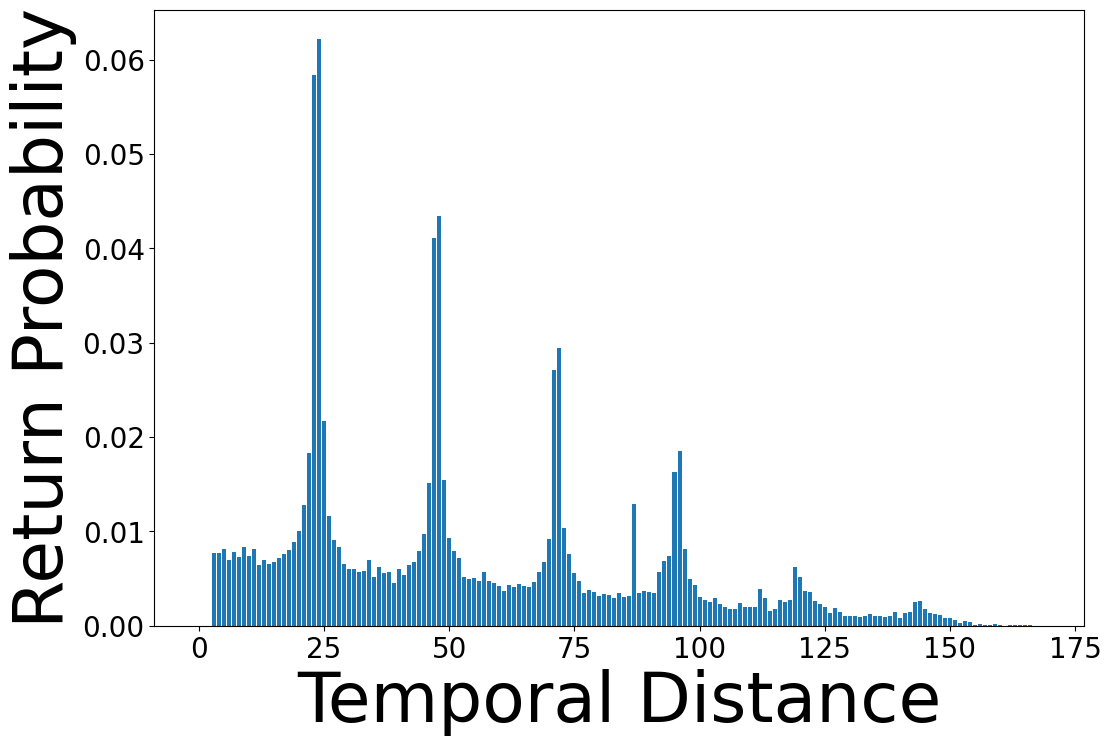}
\label{NYC_revisit_real}
}
\vspace{-0.5em}
\subfigure[MIRAGE] {
\includegraphics[width=0.3\columnwidth]{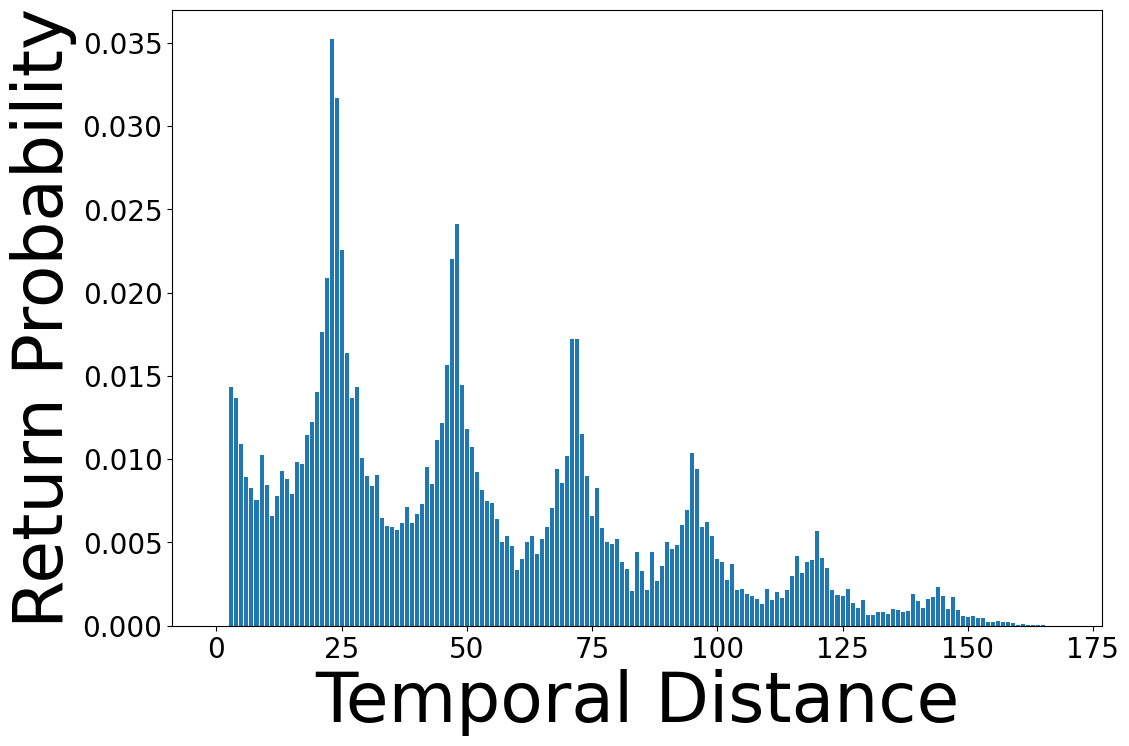}
\label{NYC_revisit_MIRAGE}
}
\vspace{-0.5em}
\subfigure[MIRAGE-noEPR] {
\includegraphics[width=0.3\columnwidth]{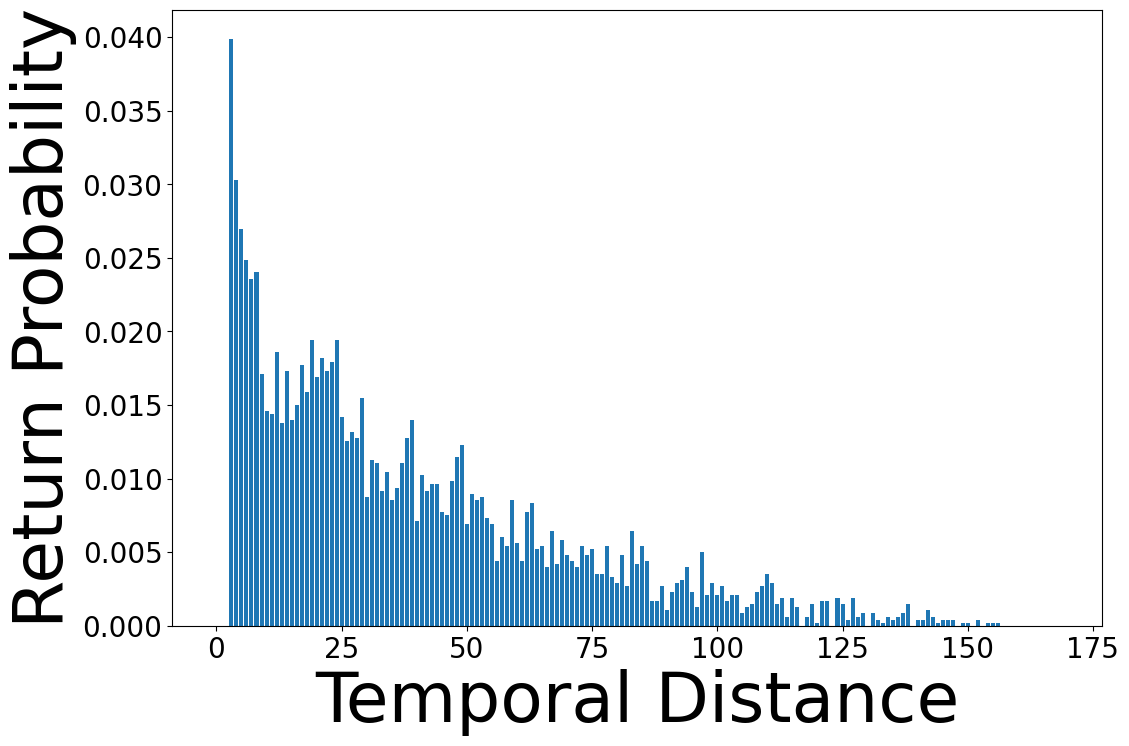}
\label{NYC_revisit_noEPR}
}
\vspace{-0.4em}
\subfigure[Semi-Markov] {
\includegraphics[width=0.3\columnwidth]{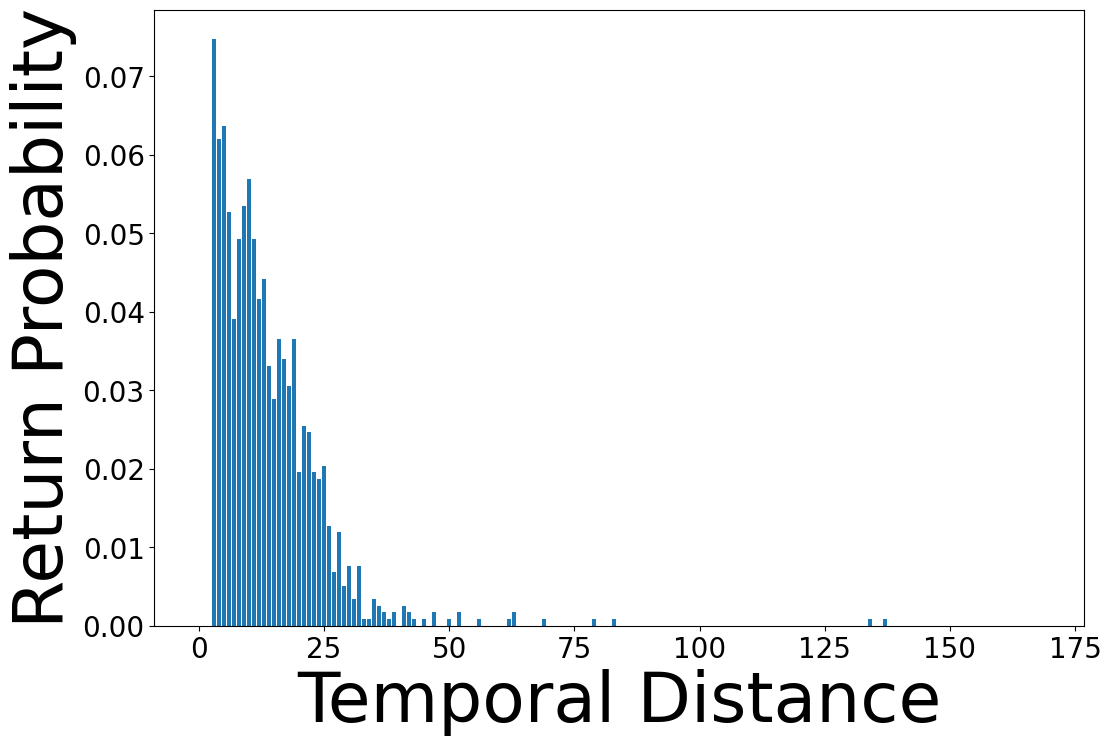}
% \label{NYC_revisit_real}
}
\vspace{-0.4em}
\subfigure[Time Geo] {
\includegraphics[width=0.3\columnwidth]{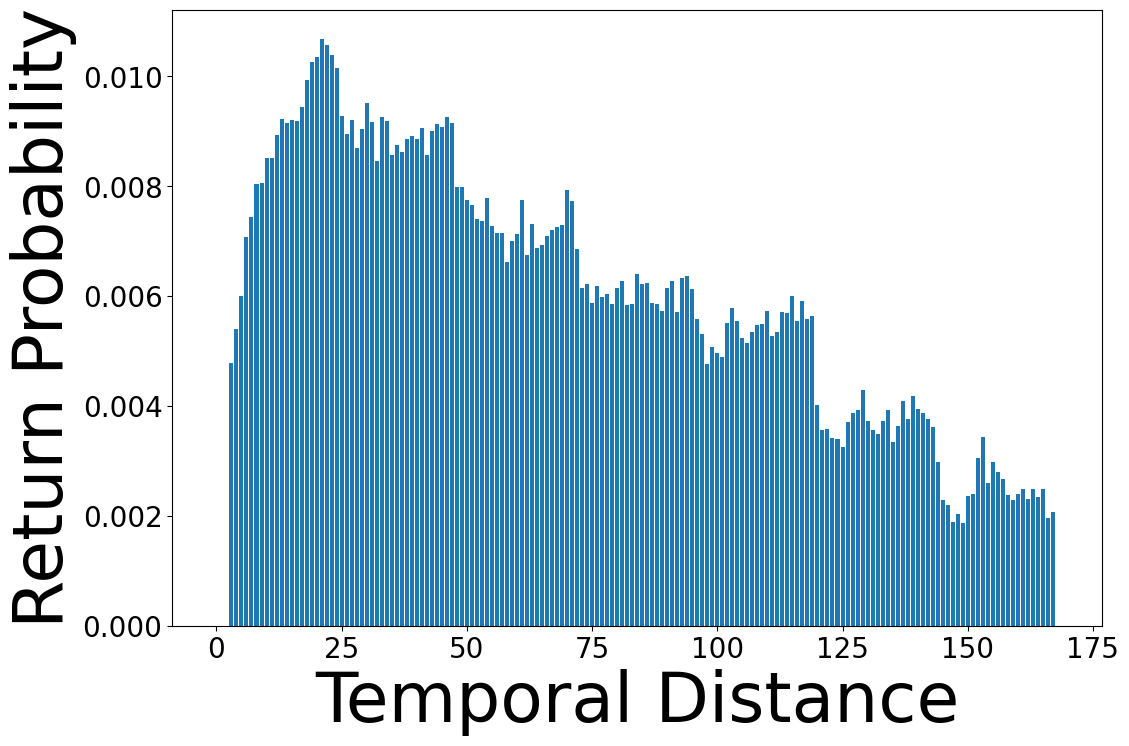}
% \label{NYC_revisit_MIRAGE}
}
\vspace{-0.4em}
\subfigure[RMTPP] {
\includegraphics[width=0.3\columnwidth]{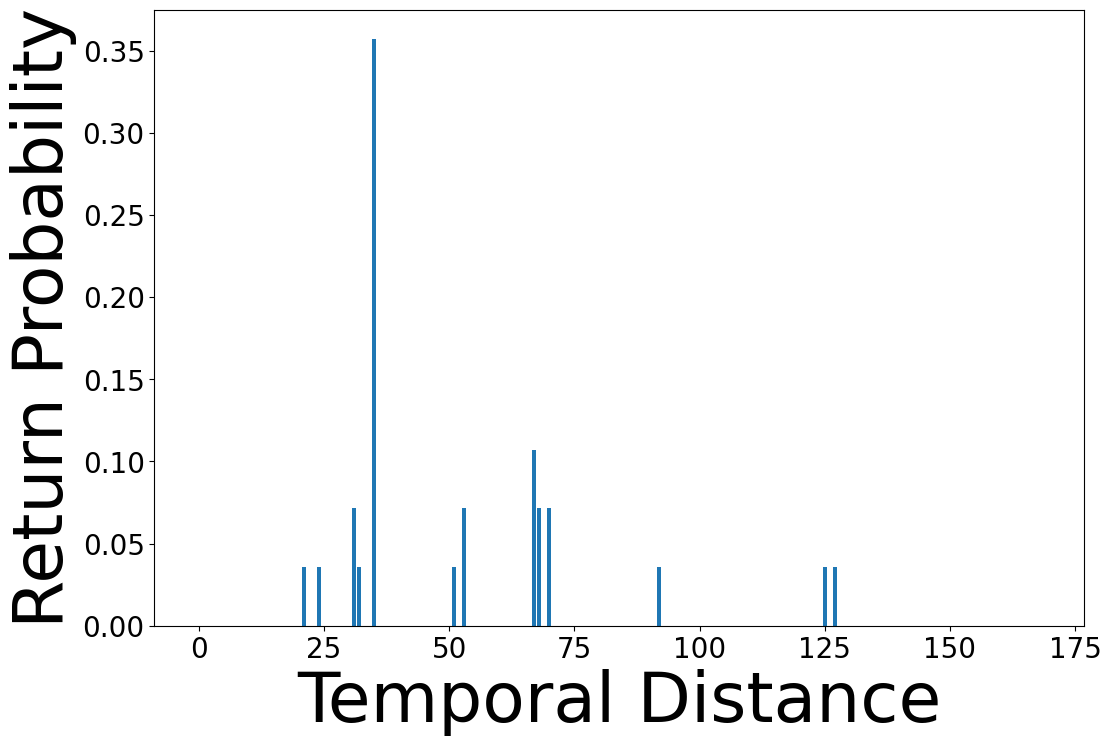}
% \label{NYC_revisit_noEPR}
}
\vspace{-0.4em}
\subfigure[ERTPP] {
\includegraphics[width=0.3\columnwidth]{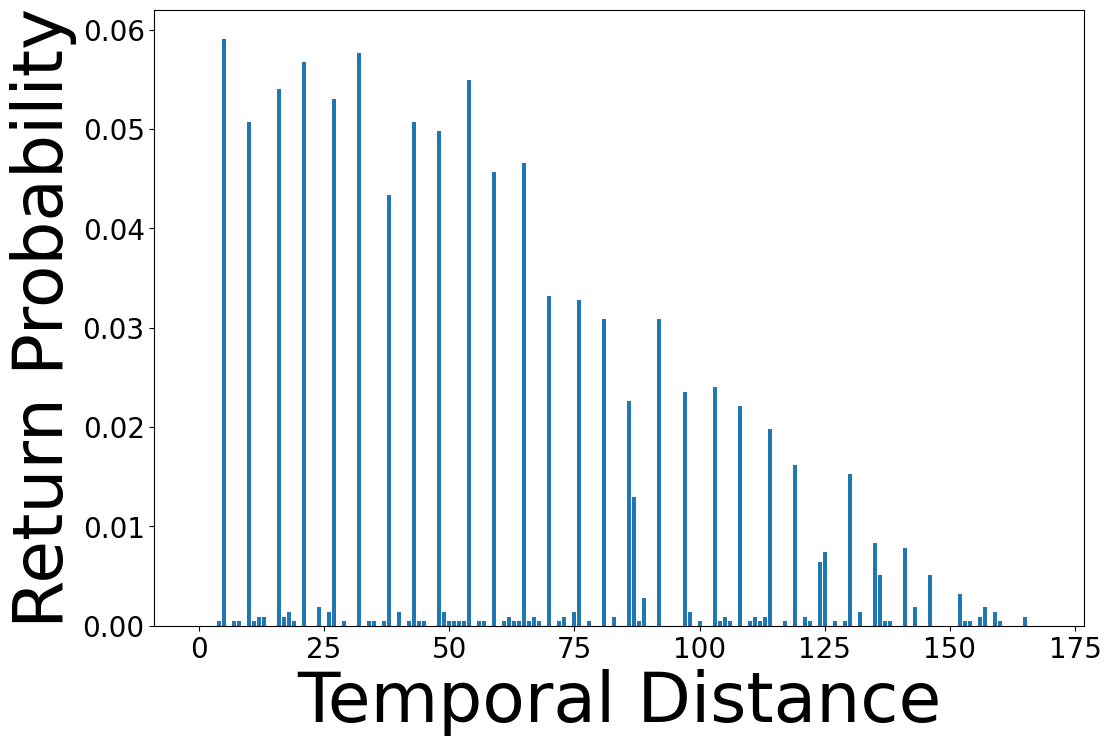}
% \label{NYC_revisit_real}
}
\vspace{-0.4em}
\subfigure[THP] {
\includegraphics[width=0.3\columnwidth]{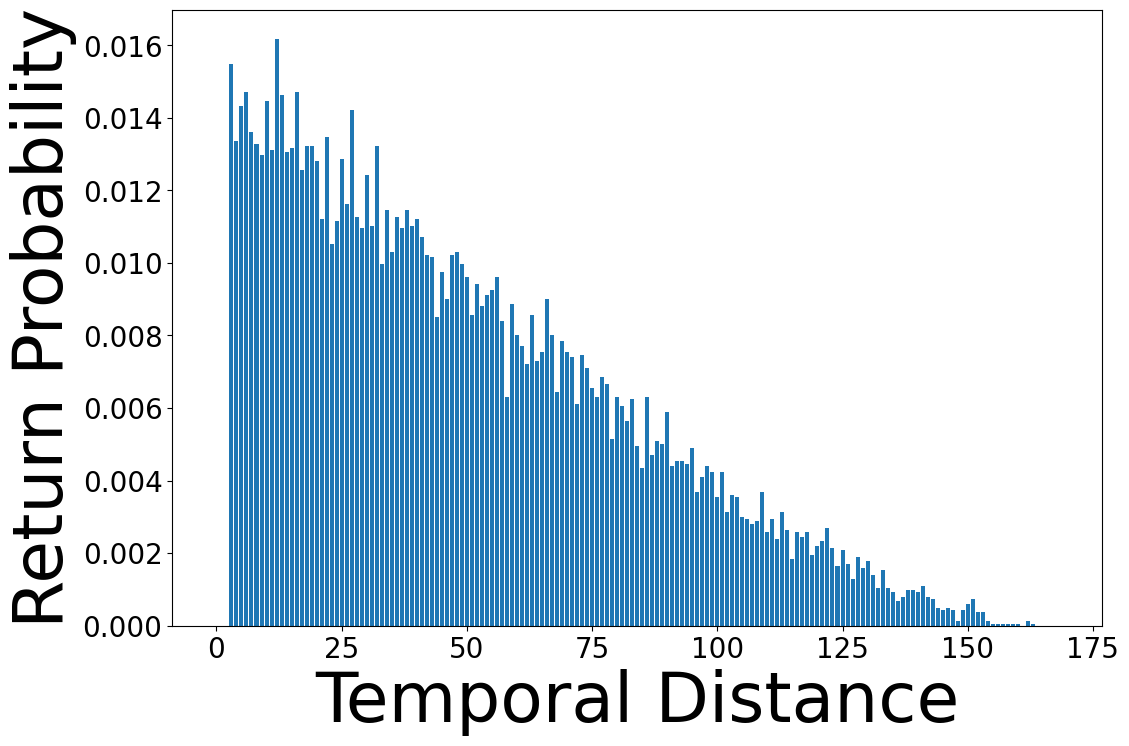}
% \label{NYC_revisit_MIRAGE}
}
\vspace{-0.4em}
\subfigure[ActSTD] {
\includegraphics[width=0.3\columnwidth]{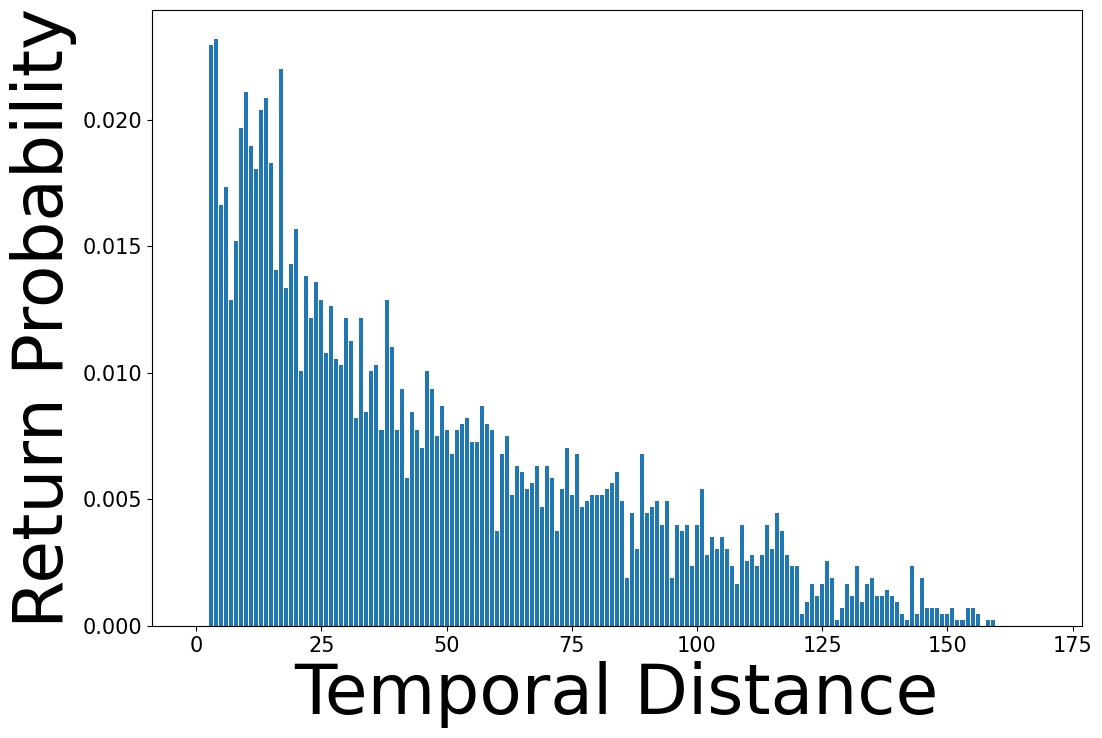}
% \label{NYC_revisit_noEPR}
}
\vspace{-0.4em}
\subfigure[LogNormMix] {
\includegraphics[width=0.3\columnwidth]{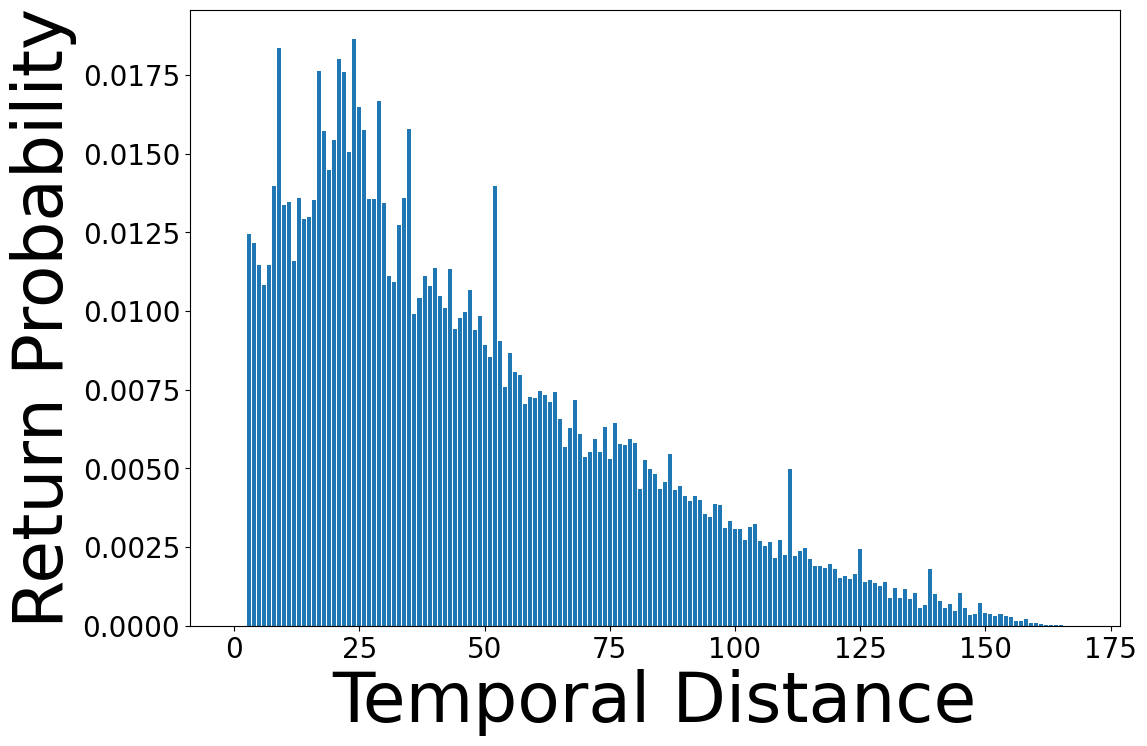}
% \label{NYC_revisit_real}
}
\vspace{-0.4em}
\subfigure[LSTM] {
\includegraphics[width=0.3\columnwidth]{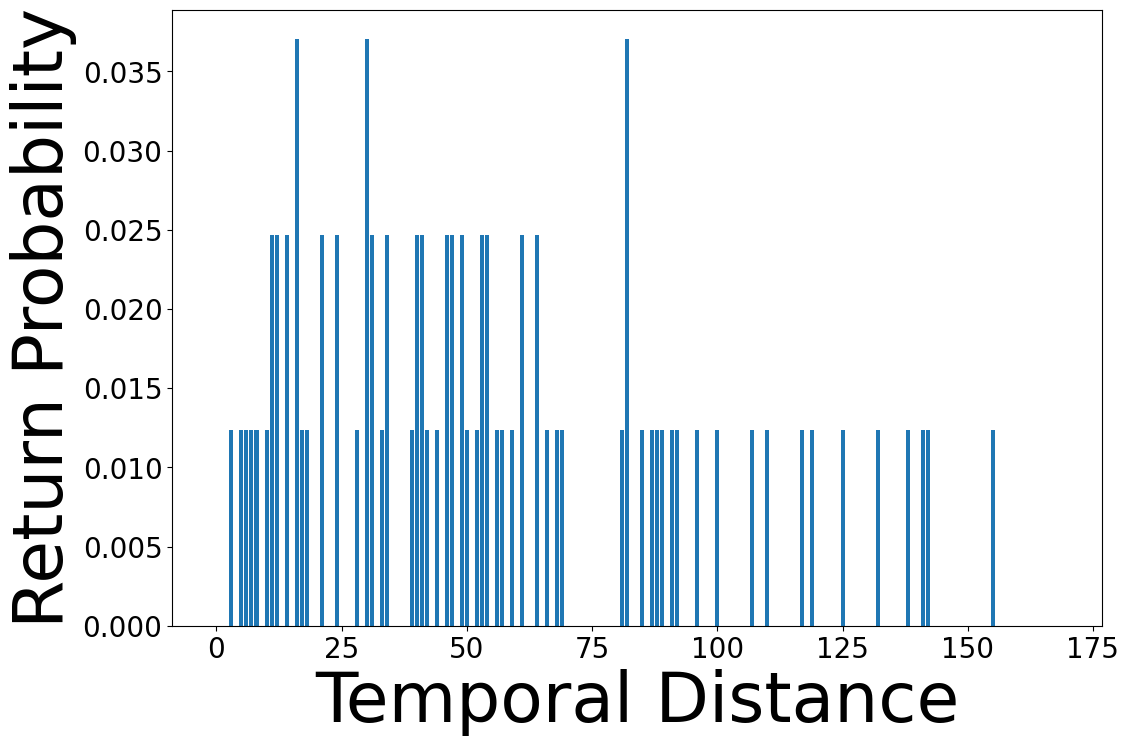}
% \label{NYC_revisit_MIRAGE}
}
\vspace{-0.4em}
\subfigure[SeqGAN] {
\includegraphics[width=0.3\columnwidth]{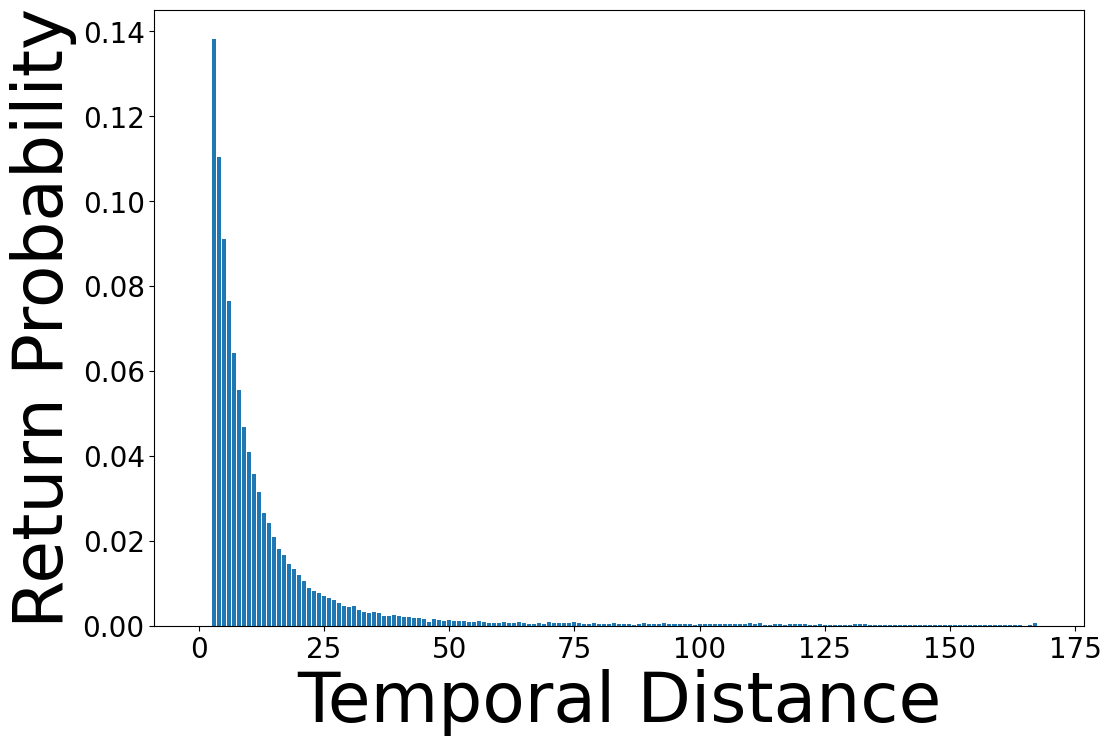}
% \label{NYC_revisit_noEPR}
}
\vspace{-0.4em}
\subfigure[MoveSim] {
\includegraphics[width=0.3\columnwidth]{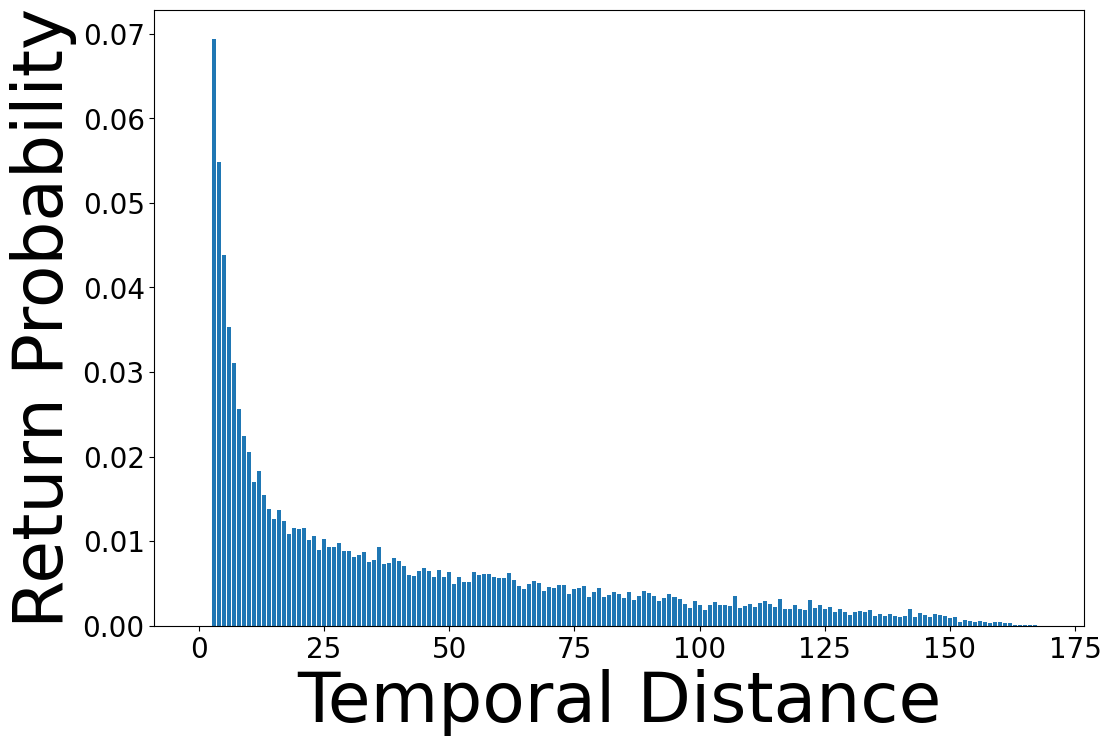}
% \label{NYC_revisit_real}
}
\vspace{-0.4em}
\subfigure[VOLUNTEER] {
\includegraphics[width=0.3\columnwidth]{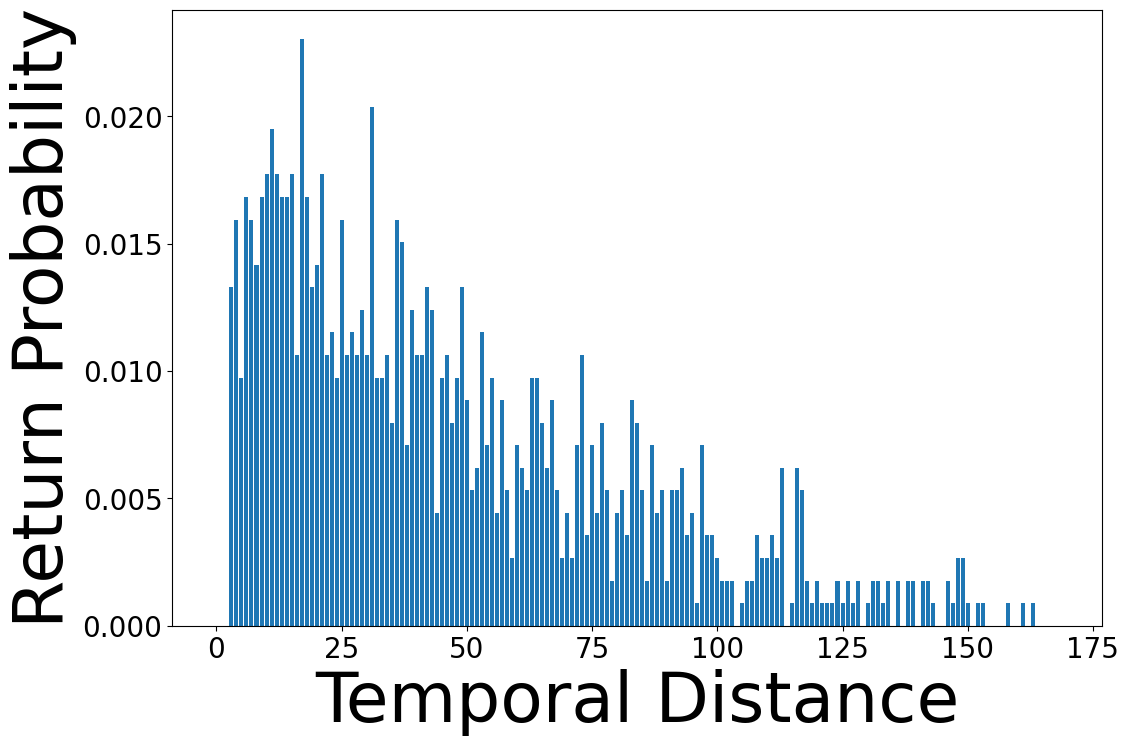}
% \label{NYC_revisit_MIRAGE}
}
\vspace{-0.4em}
\subfigure[DiffTraj] {
\includegraphics[width=0.3\columnwidth]{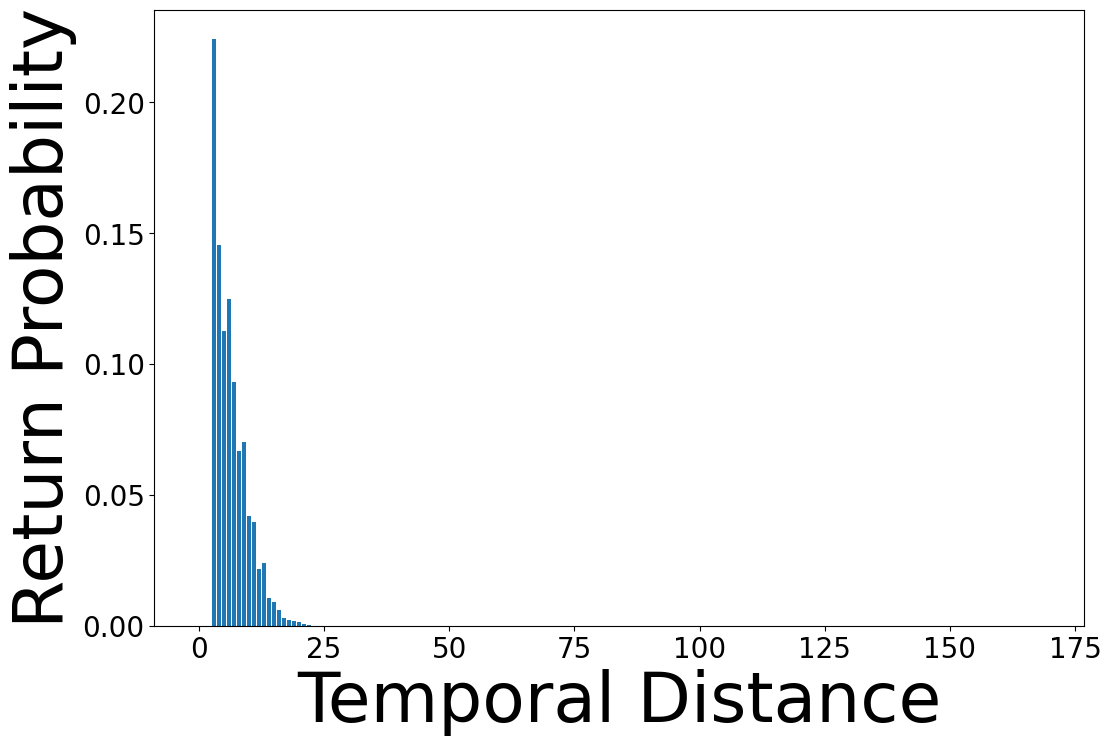}
% \label{NYC_revisit_noEPR}
}

% \vspace{-0.5em}
\caption{Returning Probability over a Week}
\vspace{-1.5em}
\label{return_prob}
\end{figure}

\begin{figure}
\centering
% \vspace{-0.5em}
\subfigure[Return Probability Distribution on Three Datasets] {
\includegraphics[width=0.43\columnwidth]{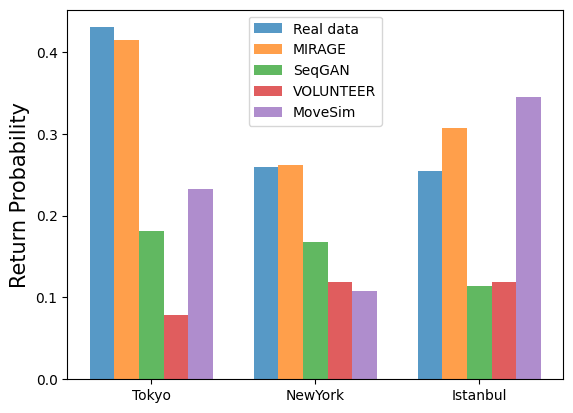}
\label{EPR_ALL}
}
% \vspace{-0.5em}
\subfigure[Return Probability Distribution over Sequence on Tokyo] {
\includegraphics[width=0.4\columnwidth]{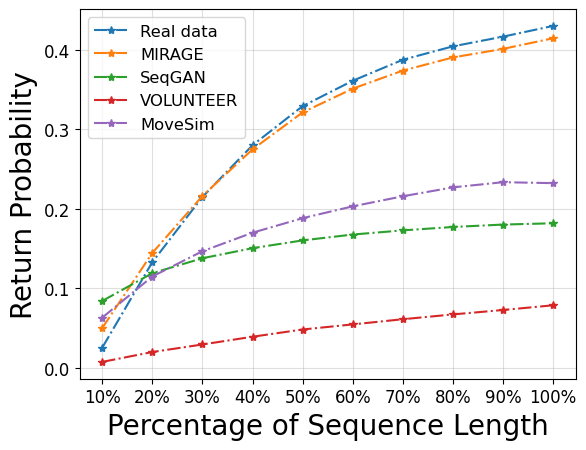}
\label{EPR_TKY}
}
% \vspace{-0.5em}
\subfigure[Return Probability Distribution over Sequence on Istanbul] {
\includegraphics[width=0.4\columnwidth]{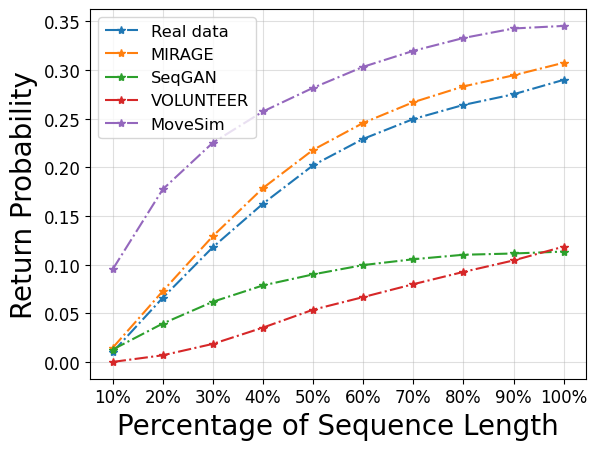}
\label{EPR_IST}
}
\vspace{-0.5em}
\subfigure[Return Probability Distribution over Sequence on NewYork] {
\includegraphics[width=0.4\columnwidth]{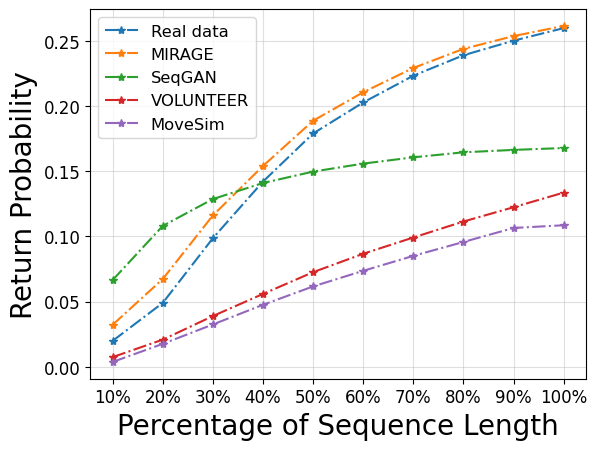}
\label{EPR_NYC}
}

\caption{Returning Probability over Sequence Length }
\vspace{-1em}
\label{return_mode_over_seq}
\end{figure}

\begin{figure}
\centering
\vspace{-0.5em}
\subfigure[Tokyo] {
\includegraphics[width=0.3\columnwidth]{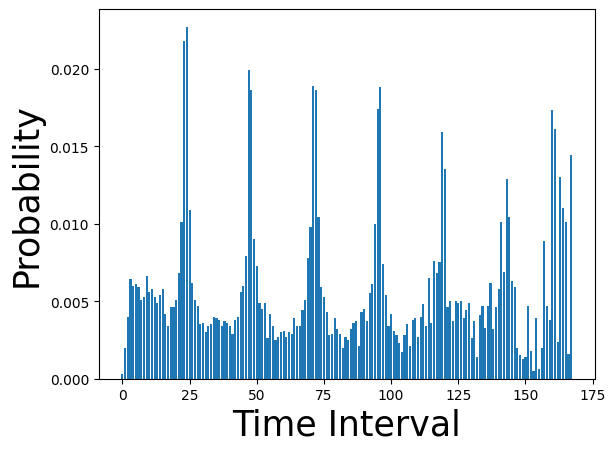}
}
% \vspace{-0.5em}
\subfigure[Istanbul] {
\includegraphics[width=0.3\columnwidth]{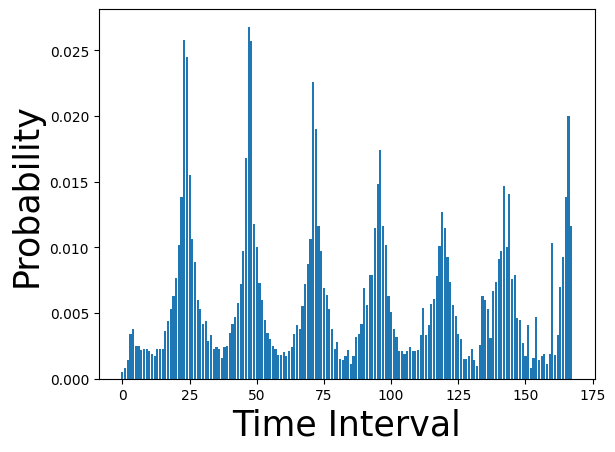}
}
% \vspace{-0.5em}
\subfigure[NewYork] {
\includegraphics[width=0.3\columnwidth]{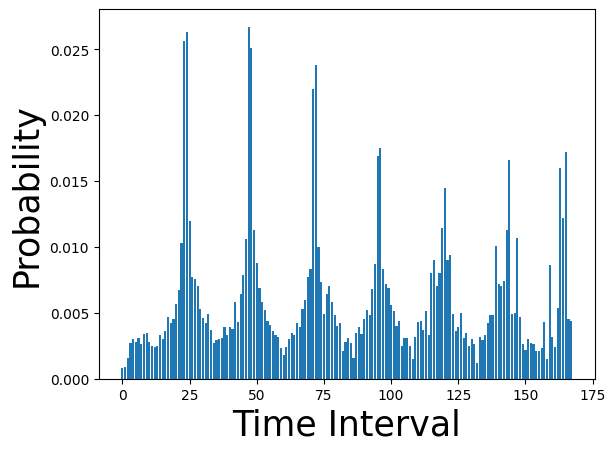}
}
\vspace{-1em}
\caption{Return probability over different time distances}
\vspace{-1em}
\label{weight_dist}
\end{figure}

In addition, to further show the utility of the EPR model, we plot the empirical returning probability of users over time, which is defined as the probability of a user returning to a location a certain period (temporal distance) after the user's first presence at the location \cite{gonzalez2008understanding, yang2020location1}. Figure \ref{return_prob} shows the plots of both real and generated data on NYC. We see that the real trajectory exhibits strong periodicity, which can be well imitated by MIRAGE but not by MIRAGE-noEPR and baselines. The superiority of MIRAGE is also validated by the return probability over sequence lengths. Figure \ref{EPR_ALL} shows the return probability of the whole trajectories of real and generated data. We selected top baselines SeqGAN, VOLUNTEER, and MoveSim for comparison. Figures \ref{EPR_TKY}, \ref{EPR_IST}, and \ref{EPR_NYC} show the detailed return probability w.r.t. the (percentage) length of trajectories of three datasets (100\% corresponds to the whole trajectory). We see that our MIRAGE is more similar to the real data in terms of return trend across different (percentage) lengths. Additionally, Figure \ref{weight_dist} shows the learnt return probability $p^*(l_{i+1}|\tau_{i+1}, return)$ over different time distances (hour granularities in a week) from each dataset. We see a clear daily return pattern, implying that our model can effectively capture the periodicity encoded in human trajectories.

\section{Conclusions and Future Works}
In this paper, by revisiting the existing human trajectory generative models, we identify their limitations in focusing on the summary statistics and distributional similarities between real and generated trajectories, which could lead to intrinsic biases in both generative model design and benchmarks of the generated trajectories. Against this background, we propose MIRAGE, a hu\underline{M}an-\underline{I}mitative t\underline{RA}jectory \underline{G}en\underline{E}rative model designed as an intensity-free neural Temporal Point Process integrating a neural Exploration and Preferential Return model to imitate the human decision-making process in trajectory generation. Meanwhile, we also propose a comprehensive task-based evaluation protocol to systematically benchmark trajectory generative models on four typical downstream tasks, integrating multiple techniques and evaluation metrics for each task, to assess the ultimate utility of the generated trajectories. The evaluation results show that MIRAGE-generated trajectory data not only achieves the best statistical and distributional similarities with 59.0-67.7\% improvement but also yields the best performance in the task-based evaluation with 10.9-33.4\% improvement.

Our future work will investigate the relationship between task performance variations and generative model design.

\begin{acks}
This project has received funding from the University of Macau (MYRG2022-00048-IOTSC), the Science and Technology Development Fund, Macau SAR (0047/2022/A1, 001/2024/SKL), and Jiangyin Hi-tech Industrial Development Zone under the Taihu Innovation Scheme (EF2025-00003-SKL-IOTSC). This work was performed in part at SICC which is supported by SKL-IOTSC, University of Macau.
\end{acks}

\bibliographystyle{ACM-Reference-Format}
\bibliography{mirage}

% \newpage
\appendix
\section{Biases in downstream tasks}
\label{sec:app_task_bias}
We show the performance variation of different methods solving the same downstream task in our task-based evaluation on MAPE in Table \ref{tab_biases}. Taking MoveSim and VOLUNTEER as examples, MoveSim is better than VOLUNTEER when benchmarking using FPMC or SASRec in this task, while we have the opposite results if benchmarking using BERT4Rec, Caser, or SRGNN.

% \section{Return Probability Visualization}
% \label{sec:weight_dist}
% Figure \ref{weight_dist_appendix} shows the learnt return probability $p^*(l_{i+1}|\tau_{i+1}, return)$ over different time distances (hour granularities in a week) from each dataset. We see a clear daily return pattern, indicating that our model can effectively capture the periodicity encoded in human trajectories.

% normalized weight of different time intervals

% \begin{figure}
% \centering
% \vspace{-0.5em}
% \subfigure[Tokyo] {
% \includegraphics[width=0.3\columnwidth]{./figures/TKY_WEIGHT.png}
% }
% % \vspace{-0.5em}
% \subfigure[Istanbul] {
% \includegraphics[width=0.3\columnwidth]{./figures/IST_WEIGHT.png}
% }
% % \vspace{-0.5em}
% \subfigure[NewYork] {
% \includegraphics[width=0.3\columnwidth]{./figures/NYC_WEIGHT.png}
% }
% \caption{Return probability over different time distances}
% \label{weight_dist_appendix}
% \end{figure}

\begin{table}
\caption{MAPE Performance of individual methods solving the NexLoc task on the IST dataset}
\label{tab_biases}
\centering
\small
\vspace{-1em}
\begin{tabular}{c|ccccc}
\hline
& FPMC & BERT4Rec & Caser & SRGNN & SASRec \\
\hline
Semi-Markov & 0.9968 & 0.9965 & 0.9960 & 0.9983 & 0.9971 \\
Time Geo & 3.5045 & 3.7099 & 3.5759 & 3.3284 & 2.9051 \\
\hline
RMTPP & 0.8838 & 0.7807 & 0.7849 & 0.8156 & 0.8250 \\
ERTPP & 0.7783 & 0.7618 & 0.7726 & 0.8007 & 0.8136 \\
THP & 0.9598 & 0.7937 & 0.7955 & 0.8205 & 0.8758 \\
ActSTD &0.8454 &0.4274 &0.5147 &0.4729 &0.6069 \\
LogNormMix & 0.8442 & 0.8523 & 0.8689 & 0.8686 & 0.8621 \\
\hline
LSTM & 0.6218 & 0.5986 & 0.6191 & 0.6707 & 0.6887 \\
SeqGAN & 0.5952 & 0.7406 & 1.7957 & 0.1857 & 0.1685 \\
MoveSim & 0.5114 & 0.5820 & 0.5215 & 0.7367 & 0.5557 \\
VOLUNTEER & 0.5279 & 0.4924 & 0.5147 & 0.5719 & 0.6019 \\
DiffTraj & 1.3272 & 1.2091 & 0.9671 & 1.0018 & 1.0399 \\
\hline
MIRAGE & 0.3111 & 0.2912 & 0.2921 & 0.3195 & 0.3089 \\
\hline
\end{tabular}
\vspace{-1.5em}
\end{table}

\section{The Complexity of MIRAGE}
\label{complexity_appendix}
Let $n$ denote the sequence length, $d$ is the GRU hidden state size, and $z$ is the latent size of user VAE. Our time decoder with intensity-free TPP and category decoder are linear layers, taking $\mathcal{O}(nd)$ time complexity and $\mathcal{O}(d)$ space complexity. Similarly, the location decoder with two modes with time complexity $\mathcal{O}(2nd)$ and space complexity $\mathcal{O}(2d)$. The GRU backbone has time and space complexities of $\mathcal{O}(nd^2)$ and $\mathcal{O}(d^2)$, respectively. For the user VAE, we consider both input and hidden size to be $d$ for simplicity, and both time and space complexities are $\mathcal{O}(d^2)$ as usually $d^2\gg dz$. In summary, MIRAGE has time complexity $\mathcal{O}(nd^2)$ and space complexity $\mathcal{O}(d^2)$.

\begin{table*}[]
\caption{Performance in the Task-Based Evaluation on MSPE}
\label{down_performance_L2}

\vspace{-1em}
\small
\begin{tabular}{l|cccc|cccc|cccc}
\hline
& \multicolumn{4}{c|}{TKY} & \multicolumn{4}{c|}{IST} & \multicolumn{4}{c}{NYC} \\
\hline
Method & LocRec & NexLoc & SemLoc & EpiSim & LocRec & NexLoc & SemLoc & EpiSim & LocRec & NexLoc & SemLoc & EpiSim \\
\hline
Semi-Markov & 0.7117 & 0.9839 & 0.2929 & 0.4125 & 0.8427 & 0.9939 & 0.2940 & 0.3378 & 0.7455 & 0.9903 & 0.2402 & 7.0165 \\
Time Geo & 0.6558 & 3.4390 & 0.0800 & 0.7302 & 14.1917 & 12.0970 & 0.2351 & 0.7822 & 1.2609 & 10.0716 & 0.1004 & 0.6650 \\
\hline
RMTPP & 0.7664 & 0.7472 & 0.1040 & 0.8849 & 0.5395 & 0.6712 & 0.2150 & 0.7432 & 0.8020 & 0.8854 & 0.1785 & 0.7773 \\
ERTPP & 0.2322 & 0.6696 & 0.0405 & 0.3146 & 0.1995 & 0.6179 & 0.2209 & 0.5942 & 0.5054 & 0.8387 & 0.5833 & 13.0105 \\
THP & 5.1891 & 5.6151 & 0.1768 & 0.1322 & 0.3651 & 0.7251 & 0.2737 & 0.9674 & 35.1497 & 16.2586 & 0.1697 & \textbf{0.4959} \\
ActSTD &0.2921 &0.4954 &0.1528 &0.1044 &0.3556 &0.3508 &0.2045 &0.2343 &0.7544 &0.6861 &0.1763 &0.9249 \\
LogNormMix & 0.2440 & 0.5112 & 0.0381 & 0.0299 & 0.2089 & 0.7387 & 0.1781 & \textbf{0.0473} & 0.5628 & 0.7577 & 0.3148 & 0.5571 \\
\hline
LSTM & 0.2653 & 0.5229 & 0.0355 & 0.0355 & 0.1955 & 0.4116 & 0.1820 & 0.1619 & 0.5268 & 0.2871 & 0.1221 & 0.4960 \\
SeqGAN & 0.2204 & 0.2290 & 0.0986 & 0.0453 & 1.2233 & 0.8872 & 0.1936 & 0.1986 & 4.8884 & 0.3213 & 0.5830 & 5.9233 \\
MoveSim & 1.2413 & 0.1249 & 0.0870 & 0.0353 & 30.6287 & 0.3623 & 0.2802 & 0.3817 & 5.2941 & 0.6145 & 0.2053 & 5.7852 \\
VOLUNTEER & 1.2227 & 0.2921 & 0.0972 & 0.2110 & 0.3791 & 0.2987 & 0.2433 & 0.1426 & 0.4571 & 0.7171 & 0.6019 & 2.9920 \\
DiffTraj & 0.5900 & 1.2704 & 0.2280 & 0.7155 & 0.5444 & 1.3134 & 0.2570 & 1.6888 & \textbf{0.4514} & 4.0397 & 0.1984 & 0.8163 \\
\hline
MIRAGE & \textbf{0.1416} & \textbf{0.1026} & \textbf{0.0222} & \textbf{0.0289} & \textbf{0.1792} & \textbf{0.0941} & \textbf{0.1235} & 0.0770 & 0.5062 & \textbf{0.0592} & \textbf{0.0147} & 0.8948 \\
\hline
\end{tabular}
\vspace{-0.5em}
\end{table*} 

\begin{table*}[]
\caption{Ablation study in the Task-Based Evaluation on MSPE}
\label{res_ablation_study_downstream_L2}
\vspace{-1em}
\small
\begin{tabular}{l|cccc|cccc|cccc}
\hline
& \multicolumn{4}{c|}{TKY} & \multicolumn{4}{c|}{IST} & \multicolumn{4}{c}{NYC} \\
\hline
Method & LocRec & NexLoc & SemLoc & EpiSim & LocRec & NexLoc & SemLoc & EpiSim & LocRec & NexLoc & SemLoc & EpiSim \\ \hline
MIRAGE-noTPP & 0.1583 & 0.1100 & 0.0397 & 0.4335 & 0.2176 & 0.2063 & 0.1631 & 0.2274 & 0.5299 & 0.3324 & 0.0184 & 3.8921 \\
MIRAGE-noEPR & 0.1517 & 0.1893 & 0.0288 & 0.0339 & 0.1879 & 0.2618 & 0.1869 & 0.0862 & 0.7906 & 0.2815 & 0.0173 & 0.9989 \\
MIRAGE-noIMI &0.2250 &0.2355 &0.0421 &0.0364 &0.2670 &0.1294 &0.1252 &\textbf{0.0729} &0.6697 &0.0810 &0.0982 &0.9606 \\
MIRAGE-TD &0.2352 &0.2295 &0.0653 &0.0344 &0.1903 &0.1385 &0.1264 &0.0997 &0.8839 &0.0649 &0.1032 &1.1433 \\\hline
MIRAGE & \textbf{0.1416} & \textbf{0.1026} & \textbf{0.0222} & \textbf{0.0289} & \textbf{0.1792} & \textbf{0.0941} & \textbf{0.1235} & 0.0770 & \textbf{0.5062} & \textbf{0.0592} & \textbf{0.0147} & \textbf{0.8948} \\
\hline
\end{tabular}
\vspace{-0.5em}

\end{table*}

\section{Baselines} \label{sec:app_baseline}
\textbf{Semi-Markov \cite{korolyuk1975semi}} model uses exponential distribution with Gamma priors for time intervals and incorporates a Dirichlet prior to construct the transition matrix for Bayesian inference. \textbf{TimeGeo \cite{jiang2016timegeo}} designs a statistical EPR model integrated with temporal information including weekly home-based tour number, dwell rate, and burst rate to further characterize temporal choices on human mobility. \textbf{RMTPP} \cite{du2016recurrent} uses Recurrent Neural Networks to jointly model the time and mark (location ID) dependency over history information. \textbf{ERTPP} \cite{ertpp} adopts distinct RNNs to independently model the timing of the next event and its associated mark. \textbf{THP} \cite{zuo2020transformer} incorporates the self-attention mechanism with the Hawkes Process to capture long-term dependencies in event sequence data. \textbf{LSTM \cite{huang2015bidirectional}} learns the sequence patterns of human trajectory to predict both the location and time of the next event in a human trajectory. \textbf{SeqGAN} \cite{yu2017seqgan} introduces reinforcement learning into the GAN model to solve the sequence generation problem. \textbf{MoveSim \cite{feng2020learning}} is a GAN-based framework that integrates physical regularities and prior knowledge of human mobility in trajectory generation. \textbf{ActSTD} \cite{yuan2022activity} enhances the dynamic modeling of individual trajectories by utilizing neural ordinary equations. \textbf{LogNormMix \cite{shchur2019intensity}} defines an intensity-free neural TPP modeling the conditional probability density distribution of trajectory events as a log-normal mixture. We extend it as marked TPP by sampling the time and location (i.e., mark) of the next event in a trajectory. \textbf{VOLUNTEER \cite{long2023practical}} incorporates a two-layer VAE model with a temporal point process to capture the characteristics of human mobility from both group and individual views. \textbf{DiffTraj \cite{zhu2024difftraj}} uses a diffusion probabilistic model for continuous location generation. We choose the nearest POI to each generated GPS coordinate as the generated POI.

\section{Algorithms for the RecLoc Task}
\label{detail_locrec}
\textbf{BPR} \cite{bpr} is a recommendation approach designed for implicit feedback. It operates by minimizing a pairwise ranking loss to learn user preferences on times effectively. \textbf{DMF} \cite{dmf} incorporates a matrix factorization model with neural network architecture to the representations of users and locations. \textbf{LightGCN} \cite{he2020lightgcn} leverages neighborhood aggregation technique to learn user and location representations on a user-location interaction graph. \textbf{MultiVAE} \cite{multivae} extends Variational Autoencoders (VAEs) to collaborative filtering for implicit feedback. \textbf{NeuMF} \cite{neumf} proposes a neural network-based collaborative filtering technique and modeling user–location interaction function with non-linearities.

% We present the details of the algorithms used for the location recommendation task. 

% \begin{itemize}[leftmargin=*]
%     \item \textbf{BPR} \cite{bpr} is a recommendation approach designed for implicit feedback. It operates by minimizing a pairwise ranking loss to learn user preferences on times effectively.
%     \item \textbf{DMF} \cite{dmf} incorporates a matrix factorization model with neural network architecture to the representations of users and locations. 
%     \item \textbf{LightGCN} \cite{he2020lightgcn} leverages neighborhood aggregation technique to learn user and location representations on a user-location interaction graph.
%     \item \textbf{MultiVAE} \cite{multivae} extends Variational Autoencoders (VAEs) to collaborative filtering for implicit feedback.
%     \item \textbf{NeuMF} \cite{neumf} proposes a neural network-based collaborative filtering technique and modeling user–location interaction function with non-linearities.
% \end{itemize}

\section{Algorithms for the Nexloc Task}
\label{detail_seqrec}
\textbf{FPMC} \cite{fpmc} uses matrix factorization techniques to estimate the personalized transition matrix of POIs in user mobility trajectories. \textbf{BERT4Rec} \cite{sun2019bert4rec} uses bidirectional self-attention network models human behavior sequences by employing a Cloze task approach. \textbf{Caser} \cite{caser} employs hierarchical and vertical CNNs to capture union-level sequential patterns and skip behaviors, enabling sequence-aware recommendation. \textbf{SRGNN} \cite{srgnn} designs a session-based recommendation model by utilizing the basic RNNs to predict the next location based on historical trajectories. \textbf{SASRec} \cite{sasrec} employs a multi-head attention mechanism to make predictions based on the historical trajectory of users.

\section{Epidemic Simulation Settings}
\label{sec:app_episim}
% \subsection{COVID-19 Simulation}
We simulate the COVID-19 spreading with the SEIR model following recent works \cite{feng2020learning,yuan2022activity,lai2020effect}. In particular, the simulation involves eight fundamental parameters: the close contact ratio ($c$), transmission period ($T$), incubation period ($T_i$), infection period ($T_f$), reproduction rate ($R_0$), transmission probability ($\beta$), infectious rate ($\alpha$), and recovery rate ($\gamma$). Table \ref{seir_parameters} shows the parameter values.

\begin{table}[]
\caption{Parameters for COVID-19 simulation ($d$: day)}
\label{seir_parameters}
\centering
\small
\vspace{-1em}
\begin{tabular}{|c|c|c|c|c|c|c|c|c|}
\hline
Parameters & $c$ & $T$ & $T_i$ & $T_f$ & $R_0$ & $\beta$ & $\alpha$ & $\gamma$ \\
\hline
Value & 0.2 & $5.8d$ & $5.2d$ & $11d$ & 2.2 & $R_0/T$ & $1/T_i$ & $1/T_f$ \\
\hline
\end{tabular}
\vspace{-1em}
\end{table}

During the simulation, we assume that infected or exposed individuals contact with $s$ susceptible individuals connected by edges in the contact network each day. The probability of two people with an edge in the contact network becoming close contact is $c$. The transmission probability $\beta$ is calculated by dividing the basic reproduction rate $R_0$ by the average duration (5.8 days) from onset to first medical visit and isolation. The infectious rate from the exposed state $\alpha$ is estimated as the reciprocal of the incubation period, which is the average time exposed but not infectious (5.2 days in \cite{lai2020effect}). The daily probability of transitioning to the removed state from infectious $\gamma$, is computed based on the average infection period (11 days in \cite{lai2020effect}). The the infection process are as follows:
\begin{equation}
\label{covid_19_equation}
    \frac{dS}{dt} = -sc\beta, \frac{dE}{dt} = sc\beta - \alpha E, \frac{dI}{dt} = \alpha E - \gamma I, \frac{dR}{dt} = \gamma I
\end{equation}
where S represents the number of susceptible individuals, E is the number of exposed individuals, I stands for the number of infectious individuals, and R denotes the number of removed individuals. We randomly select 50 individuals as exposed and label their status accordingly. Using the differential equations \ref{covid_19_equation}, we simulate the spread of the epidemic and record the daily counts of exposed, infectious, and removed individuals.

% \subsection{Influenza Simulation}
Similarly to the COVID-19 simulation, we simulate influenza spreading using the parameters suggested by \cite{brauer2019models}, where the transmission probability $\beta$ infectious rate $\alpha$ and removed rate $\gamma$ are shown in Table \ref{seir_parameters_influenza}.

\begin{table}[]
\caption{Parameters for influenza simulation}
\label{seir_parameters_influenza}
\centering
\small
\vspace{-1em}
% \begin{tabular}{|c|c|c|c|c|c|c|c|c|}
% \hline
% Parameters & $c$ & $T$ & $T_i$ & $T_f$ & $R_0$ & $\beta$ & $\alpha$ & $\gamma$ \\
% \hline
% Value & 0.2 & $1.9d$ & $5.2d$ & $4.1d$ & 1.37 & $R_0/T$ & $1/T_i$ & $1/T_f$ \\
% \hline
% \end{tabular}
\begin{tabular}{|c|c|c|c|c|c|c|c|c|}
\hline
Parameters & $c$ &$\beta$ & $\alpha$ & $\gamma$ \\
\hline
Value & 0.2 & 0.402 & 0.526 & 0.244 \\
\hline
\end{tabular}
\vspace{-1em}
\end{table}

% \begin{table}[]
% % \setlength{\tabcolsep}{0.5em}
% \caption{Parameters for influenza simulation}
% \label{seir_parameters_influenza}
% \centering
% % \begin{tabular}{|c|c|c|c|c|c|c|c|c|}
% % \hline
% % Parameters & $c$ & $T$ & $T_i$ & $T_f$ & $R_0$ & $\beta$ & $\alpha$ & $\gamma$ \\
% % \hline
% % Value & 0.2 & $1.9d$ & $5.2d$ & $4.1d$ & 1.37 & $R_0/T$ & $1/T_i$ & $1/T_f$ \\
% % \hline
% % \end{tabular}
% \begin{tabular}{|c|c|c|c|c|c|c|c|c|}
% \hline
% Parameters & $c$ &$\beta$ & $\alpha$ & $\gamma$ \\
% \hline
% Value & 0.2 & 0.402 & 0.526 & 0.244 \\
% \hline
% \end{tabular}
% \end{table}

% \section{Performance in Task-Based Evaluation on MSPE}
\section{Performance on MSPE}
\label{sec:addtional_result}
Table \ref{down_performance_L2} shows performance comparison in the task-based evaluation on MSPE. Similar to the results on MAPE, we observe that MIRAGE achieves the best performance in most cases. Compared to the best-performing baselines, our MIRAGE achieves performance improvement with 17.6 \% on average. Table \ref{res_ablation_study_downstream_L2} shows the ablation study results in the task-based evaluation on MSPE. Similar to the results on MAPE, we also observe that MIRAGE outperforms its variants MIRAGE-noTPP, MIRAGE-noEPR, MIRAGE-noIMI, and MIRAGE-TD by 41.8\%, 28.7\%, 30.0\%, and 33.3\%, respectively, further validating our key design choices.

\end{document}